\documentclass[runningheads]{llncs}
\usepackage{graphicx}
\usepackage{comment}
\usepackage{amsmath,amssymb} 
\usepackage{color}
\usepackage{booktabs}
\usepackage{caption}
\usepackage{subcaption}
\usepackage{multirow}
\usepackage[dvipsnames]{xcolor}
\usepackage{epsfig}
\usepackage{xspace}
\usepackage{wrapfig}
\usepackage{algorithm}
\usepackage{algorithmic}

\usepackage[symbol]{footmisc}

\definecolor{myblue}{RGB}{91, 155, 213}
\definecolor{mygreen}{RGB}{112, 173, 71}
\definecolor{myorange}{RGB}{237, 125, 49}

\newcommand{\tabincell}[2]{\begin{tabular}{@{}#1@{}}#2\end{tabular}}
\newcommand{\myparagraph}[1]{\vspace{3pt}\noindent{\bf #1}}
\newcommand{\myitparagraph}[1]{\vspace{3pt}\noindent{\it #1}}

\def\cityscapes{\textit{Cityscapes}\xspace}
\def\bdd{\textit{BDD100K}\xspace}
\def\mapillary{\textit{Mapillary Vistas}\xspace}

\def\upernet{{UperNet}\xspace}
\def\pspnet{{PSPNet}\xspace}
\def\deeplab{{Deeplab-v3+}\xspace}
\def\dpc{{DPC}\xspace}

\begin{document}
\pagestyle{headings}
\mainmatter

\title{Segmentations-Leak: Membership Inference Attacks and Defenses in Semantic Image Segmentation} 

\titlerunning{Membership Inference Attacks and Defenses in Semantic Segmentation}
\authorrunning{He et al.} 
\author{Anonymous ECCV submission}


%
\author{Yang He\inst{1,2} \and
Shadi Rahimian\inst{1} \and
Bernt Schiele\inst{2} \and Mario Fritz\inst{1}}
%
%
\institute{CISPA Helmholtz Center for Information Security \and
Max Planck Institute for Informatics \\\
Saarland Informatics Campus, Germany \\
\email{\{yang.he, shadi.rahimian, fritz\}@cispa.saarland, schiele@mpi-inf.mpg.de}}
\maketitle

\begin{abstract}
   Today's success of state of the art methods for semantic segmentation is driven by large datasets. Data is considered an important asset that needs to be protected, as the collection and annotation of such datasets comes at significant efforts and associated costs. In addition, visual data might contain private or sensitive information, that makes it equally unsuited for public release. Unfortunately, recent work on membership inference in the broader area of adversarial machine learning and inference attacks on machine learning models has shown that even black box classifiers leak information on the dataset that they were trained on. We show that such membership inference attacks can be successfully carried out on complex, state of the art models for semantic segmentation. In order to mitigate the associated risks, we also study a series of defenses against such membership inference attacks and find effective counter measures against the existing risks with little effect on the utility of the segmentation method.  Finally, we extensively evaluate our attacks and defenses on a range of relevant real-world datasets: Cityscapes, BDD100K, and Mapillary Vistas. \footnote{Our source code and demos are available at \\ \url{https://github.com/SSAW14/segmentation_membership_inference}.}
   
\end{abstract}

\section{Introduction}
\label{sec:label}
\vspace{0.25cm}
The availability of large datasets is playing a key role in today's state of the art computer vision methods
ranging from image classification (e.g. ImageNet~\cite{deng2009imagenet}), over semantic segmentation~\cite{cordts2016cityscapes,mapillary2017iccv,yu2018bdd100k}, to visual question answering~\cite{antol2015vqa}. 
Therefore, research and industry alike have recognized the importance of large-scale datasets~\cite{deng2009imagenet,kay2017kinetics,sun2017revisiting,zhou2017places} to push performance of computer vision algorithms. However, data collection and in particular annotation and curation of large datasets comes at a substantial cost. There are sizable efforts from the research community \cite{cordts2016cityscapes,geiger2012kitti,yu2018bdd100k}, and also industry has picked up the task of collection (e.g. \cite{mapillary2017iccv}) as well as providing annotation services such as Amazon MTurk, which in turn can be monetized and constitutes important assets to companies. 

Consequently, such assets need protection e.g. as part of intellectual property and it should be controlled which parts are made public (e.g. for research purposes) and which part remain private. Based on these datasets, high performing models are trained and then made public (e.g. as black box models) via an API or as part of a product. One might assume that the information of the training set remains contained within the trained parameters of the model and therefore remains private.
Beyond the aspect of intellectual property, data might also include private information that were captured as part of the data collection process, which are sensitive and important for safe and clean services. 

Unfortunately, recent work on membership inference attacks \cite{white_vs_black,salem2019ml,shokri2017membership} has shown that even a black box  model leaks information of the training data, \textit{aiming to infer if a particular sample was used as part of the training data or not.} Such approaches have shown high success rates on a range of \textbf{classification} tasks and have equally proven to be hard to fully prevent (= defend). While this constitutes a potential threat to the machine learning model, it can also potentially be used as a forensics technique to detect a potentially unauthorized use of data.

However, we are still missing even a basic understanding on if and how these membership attack vectors extend to semantic segmentation, which is a basic computer vision task and has broad applications~\cite{chen2018dpc_nips,he2017std2p,lin2017refinenet,long2015fully,zhao2017pyramid}. Hence, we propose and study first membership inference attacks and defenses for semantic segmentation.
To reach this goal, we design an attack pipeline based on per-patch analysis, and discover (1) not all the areas of an input are helpful to membership inference, 
 (2) structural information itself leaks membership privacy and 
(3) effective defense mechanisms exists that can reduce the effectiveness of these attacks substantially.
Accordingly, we highlight our contributions to \textbf{segmentation} task and review relevant work.

\subsection{Contributions}
Our main contributions are as follows. (1) We present the first work on membership inference attacks against semantic segmentation models under different data/model assumptions. (2) We show structural outputs of segmentation have severe risks of leaking membership. Our proposed structured loss maps achieve the best attack results. (3) We present a range of defense methods to reduce membership leakage. In the end, we show feasible solutions to protect against membership attacks. (4) Extensive comparisons and ablation studies are provided in order to shed light on the core challenges of membership inference attacks for semantic segmentation.

\newpage

\subsection{Related Work}
Recent attacks against machine learning models have drawn much attention to communities focusing on attacking model functionality (e.g., adversarial attacks~\cite{fischer2017adversarial_segmentation,moosavi2017universal,moosavi2016deepfool,oh2017adversarial,david2019disentangling_adv_robust,xie2017adversarial_segmentation}), or stealing functionality~\cite{orekondy2019knockoff} or configurations~\cite{oh2018iclr_reverse_model} of a model. In this paper, we detail the topics of data privacy and security in the following.

\myparagraph{Membership inference attack.}
Membership inference attacks have been successfully achieved in many problems and domains, varying from biomedical data \cite{backes2016member_rna}, locations \cite{pyrgelis2017knock}, purchasing records \cite{salem2019ml}, and images \cite{shokri2017membership}. 

It has been shown that machine learning models can be attacked to infer the membership status of their training data. 
Shokri et al. \cite{shokri2017membership} proposed membership inference attacks against classification models utilizing multiple shadow models to mimic behaviors of the victim model. Shadow models were trained by querying the victim model using examples with higher confidences from the victim model. Hence, a binary classifier  was trained with information from shadow models, and applied to attack the victim. Further, Salem et al. \cite{salem2019ml} demonstrated only one shadow model is enough to reach similar results rather than multiple shadow models.
They also show that underlying  distributions of data used to train shadow models and the victim can be different, which allows for attacks under relaxed assumptions. In addition,  learning free attacks were proposed, which constitutes a low-skill attack without knowledge about the model and data distribution priors.  Salem et al.~\cite{salem2019ml} proposed to directly set a threshold on the confidence scores of predictions to recognize memberships. Sablayrolles et al.~\cite{white_vs_black} set a threshold on loss values and achieved quite successful results. 
While prior work has only studied classification models so far, our contribution is to show the differences between segmentation and classification models and present the first study of attacks and defenses on semantic segmentation models based on new methods. Although the segmentation problem can be understood as pixel-wise classification, it turns out the derived information is weak and needs to be aggregated over a patch or even the full image for a successful attack. Beyond this, we propose the first dedicated attacks that fully leverage the information of the full segmentation output and hence lead to even stronger attack vectors.

\myparagraph{Privacy-preserving machine learning.} The goal of these techniques is to reduce information leakage with limited access to training data, which have been applied to deep learning \cite{abadi2016deep,shokri2015privacy_deep_learning}.
Differential privacy \cite{dwork2011differential} allows learning the statistical properties of a dataset while preserving the privacy of the individual data points in it. 
Jayaraman et al. \cite{jayaraman2019evaluating} discussed the connection between the effectiveness of differential privacy and membership inference in practice. Besides, Nasr et al. \cite{nasr2018machine_ccs} provided membership protection for a classifier by training a coupled attacker in an adversary manner. Zhang et al. \cite{zhang2018privacy_obfuscation} obfuscated training data before feeding them to the model training task, which hides the statistical properties of an original dataset by adding random noises or providing new samples.
In our work, we compare a series of defense approaches to mitigate membership leakage in semantic segmentation.

\section{Attacks against Black-box Semantic Segmentation Models}
\label{sec:attacks}
Membership inference is to attack a \textbf{victim}, aiming to determine whether a particular data point was part of the training data of the victim.
Such attacks exploit overfitting artifacts on training data~\cite{shokri2017membership,salem2019ml,white_vs_black}. Typical machine learning models tend to be overconfident on data points that were seen during the training.
Such overfitting issues lead to characteristic patterns and  distributions of confidence scores~\cite{shokri2017membership,salem2019ml} or loss values~\cite{white_vs_black} which has facilitated membership inference attacks. As a result, successful attacks against classification models can be achieved according to a \textbf{shadow} model trained by a malicious attacker, mimicking the overfitting patterns and distribution gaps.

We show how such attacks can equally be constructed against models for semantic segmentation with a specially designed pipeline and representations. While such models can be understood as pixel-wise classification, it turns out that the information that can be derived from a single pixel is rather weak. Hence, we develop a method that aggregates such information over patches and full images to arrive at stronger attacks.
We first describe our pipeline for attacking segmentation models, and then present two attack settings exploited in our study, which have different constraints during attacks. Furthermore, we discuss our evaluation methodology, and then show evaluation results.

\vspace{-1cm}
\begin{figure}[h!]
 \begin{minipage}[t]{0.478\textwidth}
 \begin{algorithm}[H]
 \scriptsize
\caption{Training an attacker}
\label{alg:membership_train} 
\textbf{Input:} $\mathcal{D}^{S}=\{(\text{X}_i,\text{Y}_i)\}_i$, \textbf{V}, $Epoch$ \\
\textbf{Output:} Per-patch attacker \textbf{A}$_\textbf{P}$ 

\begin{algorithmic}[1]
    \STATE Query $\mathcal{D}^{S}$ with \textbf{V};
    \STATE Partition $\mathcal{D}^{S}$ into $\mathcal{D}^{S}_{in}$, $\mathcal{D}^{S}_{out}$;
    \STATE Train a shadow model \textbf{S} with $\mathcal{D}^{S}_{in}$; 
    \STATE Initialize \textbf{A}$_\textbf{P}$;
    \FOR {$i=1;i\leq Epoch;i++$}
        \FOR {$j=1;j\leq |\mathcal{D}^S|;j++$}
            \STATE Crop a patch $(\hat{\text{X}}_j,\hat{\text{Y}}_j)$ from $(\text{X}_j,\text{Y}_j)$;
            \IF {(($\text{X}_j,\text{Y}_j)\in\mathcal{D}^{S}_{in}$)}
                \STATE \textbf{A}$_\textbf{P}(\textbf{S}(\hat{\text{X}}_j),\hat{\text{Y}}_j)\xrightarrow{\text{learn}}1$
            \ELSE 
                \STATE  \textbf{A}$_\textbf{P}(\textbf{S}(\hat{\text{X}}_j),\hat{\text{Y}}_j)\xrightarrow{\text{learn}}0$
            \ENDIF
        \ENDFOR    
    \ENDFOR
    \STATE return \textbf{A}$_\textbf{P}$;
\end{algorithmic}
\end{algorithm}
 \end{minipage}
 \hfill
 \begin{minipage}[t]{0.52\textwidth}
\begin{algorithm}[H]
\scriptsize
\caption{Testing (Membership Inference)}
\label{alg:membership_inference}
\textbf{Input:} Testing pair $(\text{X},\text{Y})$,  \textbf{V}, \textbf{A}$_\textbf{P}$, $N$, $\tau$ \\
\textbf{Output:} Image-level inference result $\textbf{A}$  

\begin{algorithmic}[1]
    \STATE $\textbf{A}=0$; $i=0$;
    \WHILE {$i < N$}
        \STATE Crop $(\hat{\text{X}},\hat{\text{Y}})$ from $(\text{X},\text{Y})$; // patch selection
        \IF {\text{Mean}(\textbf{V}($\hat{\text{X}})\otimes\hat{\text{Y}})>\tau$)}
            \STATE \textbf{continue;}  // reject too confident patches
        \ENDIF
        \STATE $\textbf{A} = \textbf{A} + \textbf{A}_\textbf{P}(\textbf{V}(\hat{\text{X}}),\hat{\text{Y}}) / N$; \\
        i++;
    \ENDWHILE
    \STATE return \textbf{A};
\end{algorithmic}
\end{algorithm}
 \end{minipage}
 \hfill
\end{figure}
\vspace{-1cm}

\vspace{-0.1cm}
\subsection{Methods}
\label{subsec:methods}
Our approach infers image-level membership information based on observing predictions of segmentation models and correct labels. 
In this section, we describe our membership inference pipeline based on per-patch analysis, as summarized in Algorithm~\ref{alg:membership_train} and~\ref{alg:membership_inference}.
Further, several design choices are discussed that significantly contribute to the success of the attack and help to understand the essence in attacking semantic segmentation models with structured outputs.

\myparagraph{Notation.} We define the notation used through the paper. Let $\mathcal{D}^{\{V,S\}}=\{(\text{X}_i,\text{Y}_i)\}_i$ be two datasets including images $\text{X}\in\mathcal{R}^{H\times W\times 3}$ and densely annotated GTs $\text{Y}\in\mathcal{R}^{H\times W\times C}$ with one-hot vectors, where $C$ is the number of predefined labels.
For each dataset, we partition it into two parts for proving different membership status, i.e., $\mathcal{D}^{\{V,S\}}=\mathcal{D}_{in}^{\{V,S\}}\cup \mathcal{D}_{out}^{\{V,S\}}$.
The victim model, which is trained on $\mathcal{D}_{in}^{V}$ and we aim for attacking, is denoted as \textbf{V}.
To achieve attacks, we build a shadow semantic segmentation model \textbf{S} with $\mathcal{D}_{in}^S$, for training an attacker.
Let $\text{P}$ be the posterior of a segmentation output, i.e., $\text{P}=\textbf{S}(\text{X})$ or $\textbf{V}(\text{X})$, depending on the stages of membership inference. Our per-patch attacker is denoted as \textbf{A}$_{\textbf{P}}(\text{P} , \text{Y})$, taking $\text{P}$ and $\text{Y}$ as the inputs and outputs a binary classification score for membership status. Finally, the image-level attacker is denoted as \textbf{A}.

\myparagraph{Training.} Our method is built upon a per-patch attacker $\textbf{A}_\textbf{P}$, as described in Algorithm~\ref{alg:membership_train}. In line with previous work on membership inference~\cite{shokri2017membership,salem2019ml}, we construct a shadow model {\bf S} that is to some extent similar to {\bf V} and therefore is expected to exhibit similar behaviour and artifacts w.r.t. membership. In addition, {\bf S} aims to capture semantic relations and dependencies between different classes in structured outputs and provide training data to the patch classifier with known membership labels.
We prepare a dataset $\mathcal{D}^{S}$ with the same label space to $\mathcal{D}^{V}$, and then {\bf S} is trained on $\mathcal{D}_{in}^{V}$. The exact assumptions of our knowledge on {\bf V} that inform the construction of {\bf S} are detailed in~\ref{sec:settings}.

\myitparagraph{Construction of per-patch attacker.} 
\textbf{S} provides training data for the per-patch attacker \textbf{A}$_{\textbf{P}}$, as we have complete membership information of \textbf{S}. This allows us to train \textbf{A}$_\textbf{P}$ by achieving the binary In/Out classification on the data pairs from $\mathcal{D}_{in}^{S}$ and $\mathcal{D}_{out}^{S}$.
\textbf{A}$_{\textbf{P}}$ can be any architecture taking image-like data as inputs, and we discuss different data representations to train it as follows.

\myitparagraph{Data representation.} 
We apply two representations of a data pair $(\text{X},\text{Y}
)$ over segmentation models, as the inputs of a per-patch attacker. In other words, we train a classifier to compare the differences between P={\bf S}(\text{X}) and Y to determine the membership status of $(\text{X},\text{Y}
)$ in training the classifiers.

1. {\it  Concatenation.} We concatenate P and Y over the channel dimension, leading to a representation with size $H\times W\times 2C$.

2. {\it  Structured loss map.} The structured loss map ($\text{SLM}\in\mathbf{R}^{H\times W\times 1}$) computes the cross-entropy loss values at all the locations ($i$, $j$), where $\text{SLM}(i,j) = -\sum_{c=1}^{C}\text{Y}(i,j,c)\cdot \log(\text{P}(i,j,c))$. 
Previous work~\cite{white_vs_black} shows the success of applying a threshold on the loss value of an image pair for image classification, and this method can be easily applied to semantic segmentation.
Despite this, we show keeping structures of loss maps is still crucial to the success of attacking semantic segmentation.

\myparagraph{Testing.} Given a data pair $(\text{X},\text{Y}
)$ to determine if it was used to train $\textbf{V}$ (i.e., $(\text{X},\text{Y}
)\in\mathcal{D}_{in}^{V}$), we are able to crop a patch $(\hat{\text{X}},\hat{\text{Y}})$ from the pair, and thus the inference result is \textbf{A}$_{\textbf{P}}(\textbf{V}(\hat{\text{X}}),\hat{\text{Y}})$ according to the representation used in the training.
In order to further amplify the attack, we aggregate the information of the per-patch attack on an image-level, therefore, the final inference result is calculated by

\begin{equation}
  \textbf{A} = \frac{1}{N}\sum_{i=1}^{N} \textbf{A}_\textbf{P}(\textbf{V}(\hat{\text{X}}^i) , \hat{\text{Y}}^i),
  \label{eq:image_level_attacker}
\end{equation}
where ($\hat{\text{X}}^i$, $\hat{\text{Y}}^i$) is the $i$-th cropped patch from $(\hat{\text{X}},\hat{\text{Y}})$.

\myitparagraph{Selection of patches.} As our method is based on scoring each patch, the selection of patches plays an important role in obtaining stronger attacks. Besides, it also helps us to understand which patches are particularly important for determine the membership status of an example. Therefore, we study the influence of different patch selection schemes with the following choices:

1. {\it Sliding windows.} We crop patches on a regular grid with a fixed step size.

2. {\it Random locations.} We sample patches uniformly across the image. 

3. {\it Random locations with rejection.} We emphasize the importance of different patches for recognizing membership is not alike, therefore, this scheme aims to reject patches, which do not contribute to final results or even provide misleading information. We observe the patches with very strong confidences or very small loss should be omitted. For example, road area counts for most pixels of an image and are segmented very well, therefore, this scheme tends to select the bordering areas between a road and other classes, instead of the center of a road.

As summarized in Algorithm~\ref{alg:membership_inference}, we construct image-level membership inference attacks according to per-patch attacks, allowing us to leverage distinct patches for successful attacks. Our pipeline is flexible  to image sizes and aspect ratios if different image sizes exist in a dataset and even cross multiple datasets. 

\vspace{-0.1cm}
\subsection{Attack Settings}\label{sec:settings}
In our method, we train a shadow segmentation model $\textbf{S}$ and an attacker $\textbf{A}$ for attacking a victim segmentation model $\textbf{V}$.
Our two attack settings differ in the knowledge on  data distribution and model selection for training $\textbf{V}$ and $\textbf{S}$.

\myparagraph{Data \& model dependent attacks:}  This attack assumes that the victims model can be queried at training time of an attacker. Besides, this setting allows to train a shadow model with the same architecture to the victim.
Specifically, $\textbf{S}$ and $\textbf{V}$ have the same learning protocol and post-processing techniques during inference. Further, this attack assumes the data distributions of $\mathcal{D}^{V}$ and $\mathcal{D}^{S}$ are also identical, which comes from the same database. Last, query with a victim model is allowed to split $\mathcal{D}^{S}$ into $\mathcal{D}^{S}_{in}$ and $\mathcal{D}^{S}_{out}$, as listed in the $1$-st line of Algorithm~\ref{alg:membership_train}, that we use the examples with stronger confidences to build $\mathcal{D}^{S}_{in}$.

\myparagraph{Data \& model independent attacks:} For this attack, we only know the victim model's functionality and a defined label space. There is no query process for constructing training set for $\textbf{S}$, instead, $\textbf{S}$ is able to be trained with a dataset of the different distribution, which leads to a cheaper and more practical attack. Furthermore, the model configuration and training protocol of the victim are unknown. The goal of the shadow model is to capture the membership status for each example, and provide training data for attack model $\textbf{A}$. 
Particularly, model and data distribution are completely different to victims, even there is no query process, which might be detected on the server. Therefore, we highlight the severity of information leakage in this simplified attack.

\vspace{-0.2cm}
\subsection{Evaluation Methodology}
We evaluate the performance of membership inference attacks with \textbf{precision-recall} curves and receiver operating characteristic (\textbf{ROC}) curves.
We regard the images used during training as positive examples, and negatives if not.
Therefore, given a testing set with $M$ image pairs used to train a model and $N$ pairs not used, random guess with probability 0.5/0.5 for both classes is able to achieve precision $\frac{M}{M+N}$ and recall 0.5.
We set different thresholds in a classifier and compare its precision-recall curve to the random guess performance, to observe if attacks are successful.
Similarly, we draw the random guess behavior in a ROC curve, which is the diagonal of a plot. 
Furthermore, to compare different attacks quantitatively, we apply maximum \textbf{F-score} ($\frac{2\cdot \text{precison}\cdot \text{recall}}{\text{precison} + \text{recall}}$) in precision-recall curves and \textbf{AUC-score} in ROC curves to evaluate attack performance.
Last, our method is based on per-patch attacks, therefore, we employ the same metrics for per-patch evaluation, to help us understand and compare different attacks, as well as defense methods in section~\ref{sec:defenses}, exhaustively.

\vspace{-0.2cm}
\subsection{Evaluation Results}
\label{subsec:exp}
\myparagraph{Data and architectures.}
We conduct the experiments on street scene semantic segmentation between various datasets, including \cityscapes \cite{cordts2016cityscapes}, \bdd \cite{yu2018bdd100k} and \mapillary \cite{mapillary2017iccv}, which are captured in different countries under diverse weathers and image qualities, providing multiple domains.
Besides, we apply \pspnet~\cite{zhao2017pyramid}, \upernet~\cite{xiao2018upernet}, \deeplab~\cite{chen2018deeplab_v3plus} and \dpc~\cite{chen2018dpc_nips} as our segmentation models.
For per-patch attacker \textbf{A}$_{\textbf{P}}$, we train a ResNet-50 \cite{he2016resnet} from scratch, allowing us to visualize the regions contributing to the recognition of membership for an example by class activation mapping~\cite{zhou2016cam_cvpr}. In detail, the size of inputs for ResNet-50 is 90$\times$90 in spatial, corresponding to 713$\times$713 image patches. 

\myparagraph{Comparison methods.}
To demonstrate the effectiveness of specific considerations for segmentation models,
we compare our pipeline to previous attackers for classification models~\cite{white_vs_black,salem2019ml}.
For~\cite{salem2019ml}, we adapt their shadow model based
attacker, by regarding each location as a classification problem. We also test their learning-free attacker by only considering the mean of confidence scores of a prediction. Besides, we compare the proposed method with~\cite{white_vs_black}, which employs a threshold on the loss value.

\begin{figure}[!t]
\begin{center}
\begin{tabular}{@{}l@{}l@{}l@{}l}
   \includegraphics[width=0.2125\linewidth]{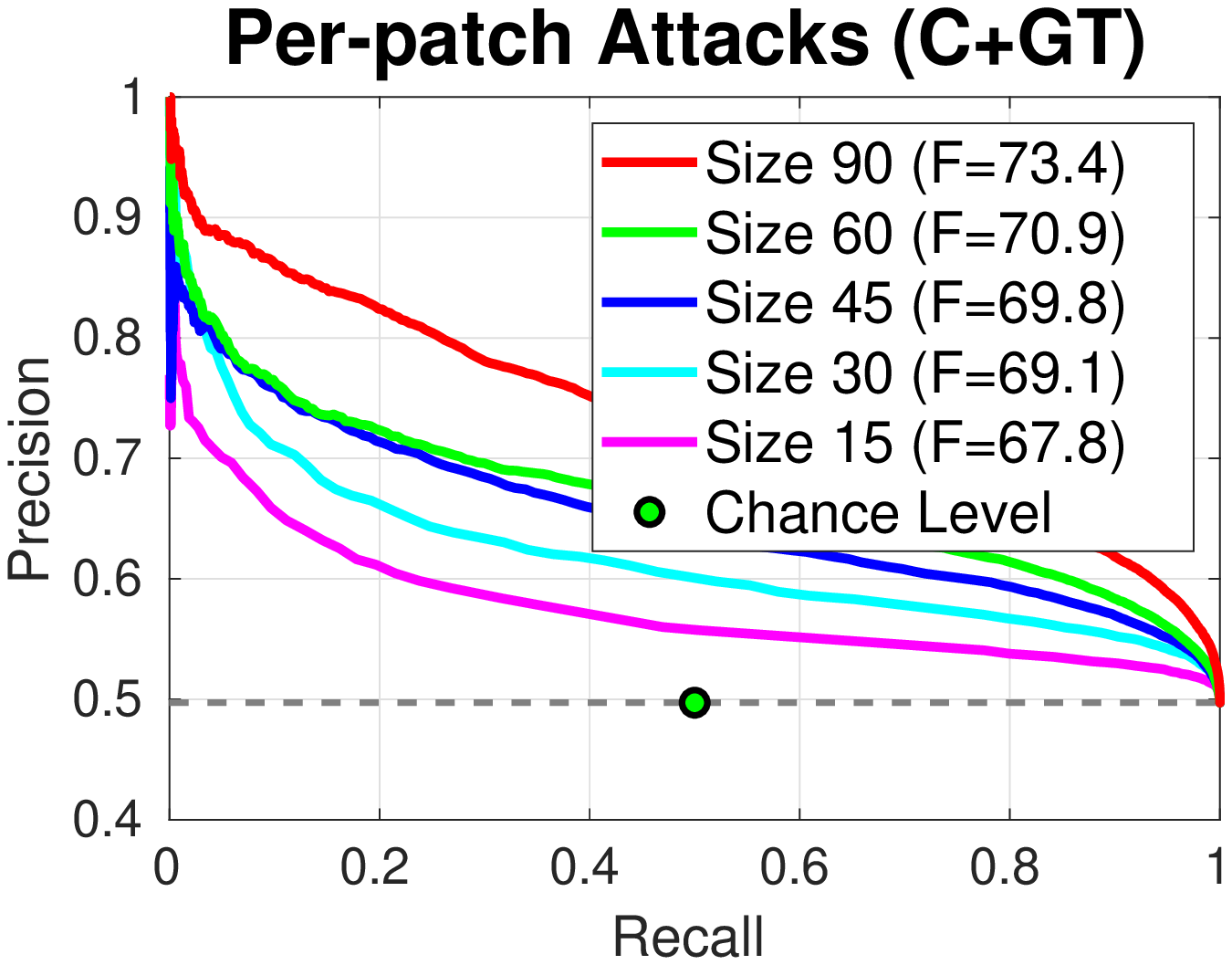} &
   \includegraphics[width=0.2125\linewidth]{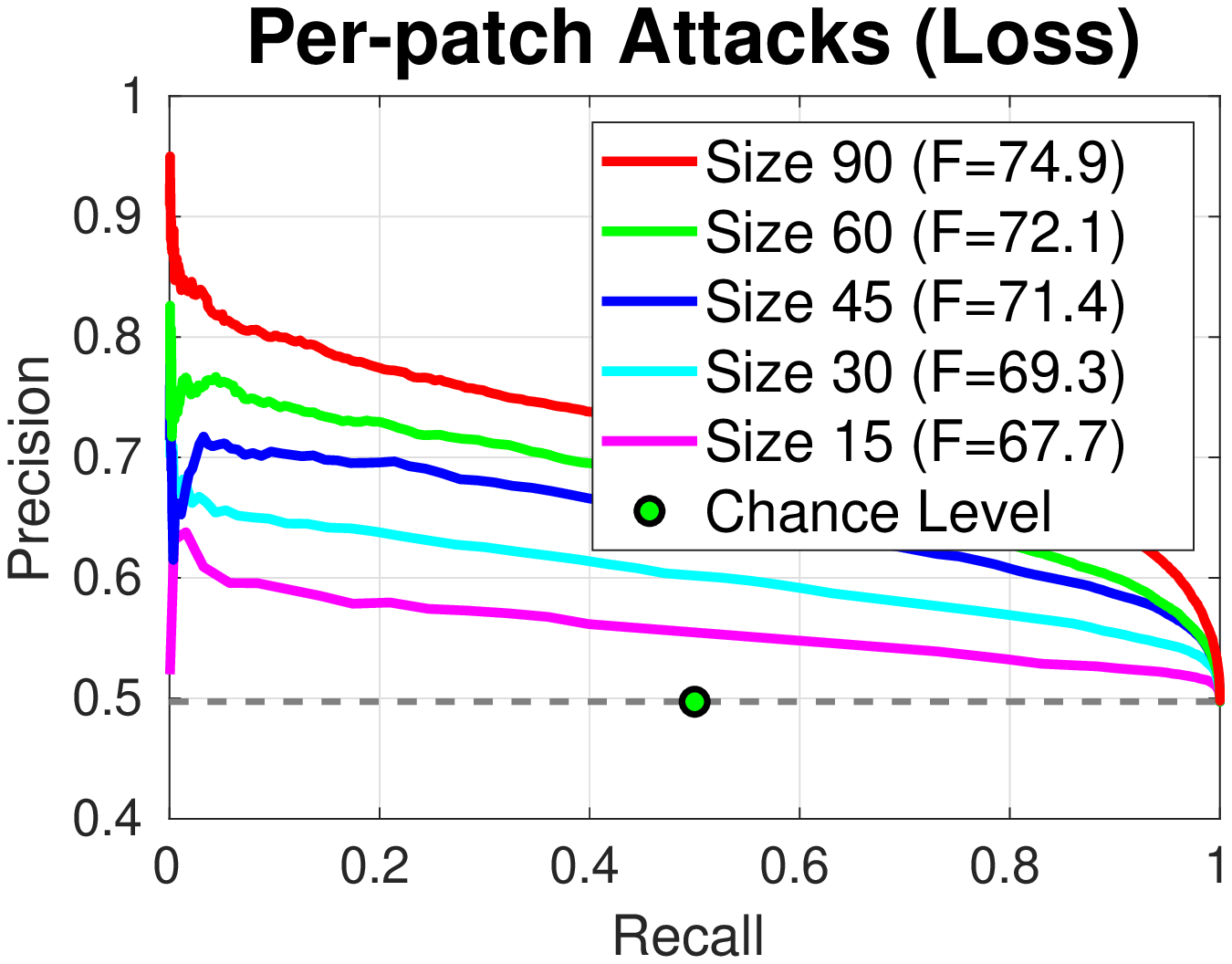} &
   \includegraphics[width=0.225\linewidth]{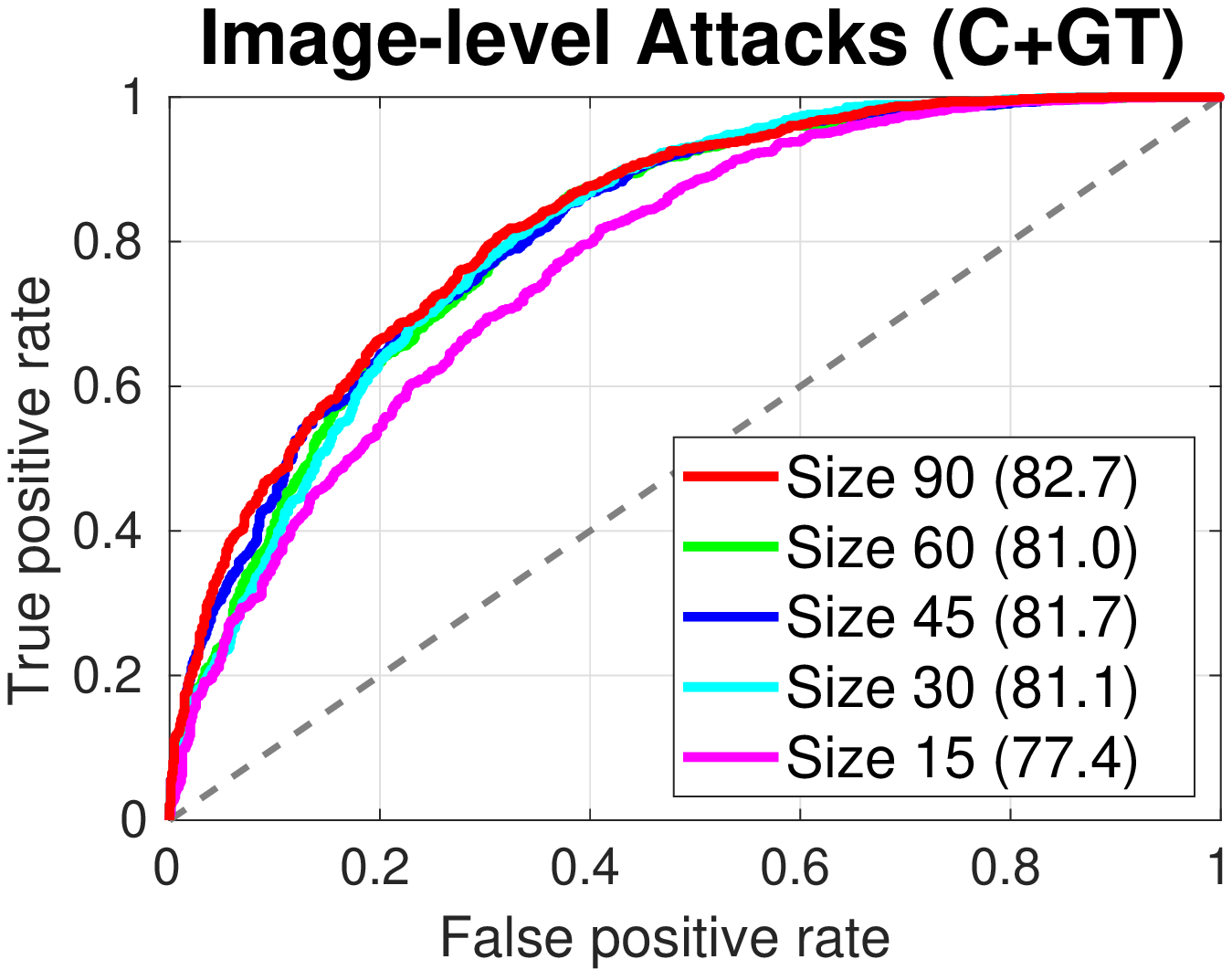} &
   \includegraphics[width=0.225\linewidth]{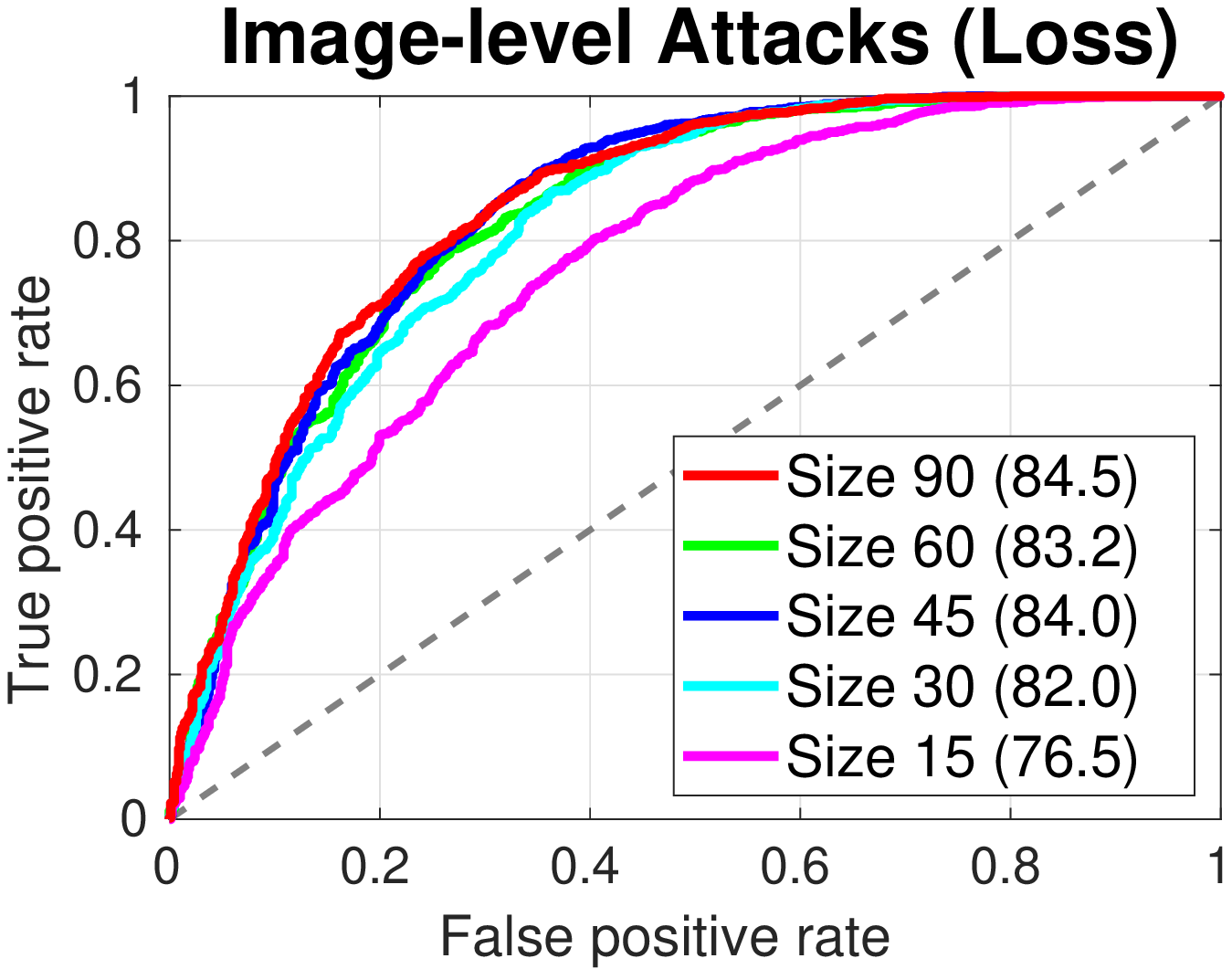} \\
   \end{tabular}
   \end{center}
   \vspace{-0.4cm}
   \caption{Evaluation of the {\bf importance of spatial structures} for PSP$\xrightarrow{}$PSP, starting from our final model (Size 90).
   \vspace{-0.3cm}   
   }
\label{fig:ablation_patch}
\end{figure}

\myparagraph{Setup for data \& model dependent attacks.} For dependent attacks, we conduct experiments with \cityscapes and \pspnet (a.k.a. PSP$\xrightarrow{}$PSP). We split \cityscapes into four parts, i.e., $\mathcal{D}_{in}^{V}$, $\mathcal{D}_{out}^{V}$, $\mathcal{D}_{in}^{S}$ and $\mathcal{D}_{out}^{V}$, where the sizes of those sets are as follows: $|\mathcal{D}_{in}^{V}|=$1488, $|\mathcal{D}_{out}^{V}|=$912, $|\mathcal{D}_{in}^{S}|=$555 and $|\mathcal{D}_{out}^{S}|=$520. We train a victim model from ImageNet~\cite{deng2009imagenet} pretrained models and lead to 59.88 mean IoU (mIoU) for segmentation. For evaluation of per-patch attacks, we sample 29760 patches from $\mathcal{D}_{in}^{V}$ and 30096 patches from $\mathcal{D}_{out}^{V}$. Therefore, this setting leads to $62\%$ and $49.7\%$ precision for image-level and per-patch attacks in random guess. The resulting F-scores for image-level and per-patch attacks are 55.36\% and 49.85\% respectively, which are drawn in Fig.~\ref{fig:ablation_patch}.

\begin{table}[t]
\tiny
\caption{Data and model descriptions of victim and shadow models for independent attacks.
}
 \centering
   \begin{tabular}{lcccc}
   \toprule
     Dataset & Model  & Backbone & In / Out \\
    \cmidrule(lr){1-1}\cmidrule(lr){2-4}
    \cityscapes (Victim) & \tabincell{c}{\pspnet \cite{zhao2017pyramid} \\ \upernet \cite{xiao2018upernet}}    & ResNet-101 \cite{he2016resnet} & 2975 / 500  \\ 
    \cmidrule(lr){1-1}\cmidrule(lr){2-4}
    \tabincell{c}{\bdd (Shadow) \\ \textit{Mapillary} (Shadow)} & \tabincell{c}{\deeplab \cite{chen2018deeplab_v3plus} \\ \dpc \cite{chen2018dpc_nips}}    & Xception-71 \cite{szegedy2016xception} & \tabincell{c}{4k / (3k+1k) \\ 10k / (8k+2k)}  \\ 
    \bottomrule
\end{tabular}
\vspace{-0.45cm}
\label{tab:data_model_overview}
\end{table}

\begin{table*}[t]
\caption{Comparison of different attackers (in \%). We compare our attackers to previous methods, including the learning-based attacker~\cite{salem2019ml}$^*$ and learning-free attackers by applying a threshold on a confidence score~\cite{salem2019ml}$^+$ or a loss value \cite{white_vs_black}.  ``$\xrightarrow{}$" means the attacks with a shadow model of the left, and the victims are the right, which can be \pspnet~\cite{zhao2017pyramid}, \upernet~\cite{xiao2018upernet}, \dpc~\cite{chen2018dpc_nips} or \deeplab~\cite{chen2018deeplab_v3plus}. } 
\tiny
\setlength{\tabcolsep}{5pt}
 \centering
   \begin{tabular}{lcccccccccc}
   \toprule
    \multirow{3}{*}{{\scriptsize Methods}}  & \multicolumn{2}{c}{{\scriptsize Dependent}} & \multicolumn{8}{c}{{\scriptsize Independent}} \\
    \cmidrule(lr){2-3}\cmidrule(lr){4-11}
     & \multicolumn{2}{c}{{\scriptsize \cite{zhao2017pyramid}$\xrightarrow{}$\cite{zhao2017pyramid}}} & \multicolumn{2}{c}{{\scriptsize \cite{chen2018deeplab_v3plus}$\xrightarrow{}$\cite{zhao2017pyramid}}} & \multicolumn{2}{c}{{\scriptsize \cite{chen2018deeplab_v3plus}$\xrightarrow{}$\cite{xiao2018upernet}}} & \multicolumn{2}{c}{{\scriptsize \cite{chen2018dpc_nips}$\xrightarrow{}$\cite{zhao2017pyramid}}} & \multicolumn{2}{c}{{\scriptsize \cite{chen2018dpc_nips}$\xrightarrow{}$\cite{xiao2018upernet}}} \\
    \cmidrule(lr){2-3}\cmidrule(lr){4-11}
     & {\scriptsize F}  & {\scriptsize AUC} & {\scriptsize F}  & {\scriptsize AUC} & {\scriptsize F}  & {\scriptsize AUC}  & {\scriptsize F}  & {\scriptsize AUC} & {\scriptsize F}  & {\scriptsize AUC} \\
     \cmidrule(lr){1-1}\cmidrule(lr){2-3}\cmidrule(lr){4-11}
    Adapted \cite{salem2019ml}$^*$ & 77.2  & 67.2 & 92.4  & 63.5 & 92.3  & 62.6   & --  & --  & --  & --  \\
    Adapted \cite{salem2019ml}$^+$ & 77.4 & 62.0 & 92.3 & 63.4 & 92.3 & 59.2 & 92.3 & 63.4 & 92.3 & 59.2 \\
    Adapted \cite{white_vs_black} & 82.2 & 74.9 & 94.4 & 81.4 & 93.0 & 72.4 & \textcolor{red}{\textbf{94.4}} & \textcolor{red}{\textbf{81.4}} & 93.0 & 72.4 \\
    Ours (C+GT, Full)  & 80.6 & 81.2 & 94.5 & 85.0 & 92.8 & 71.8 & 93.2 & 73.5 & 92.6 & 68.8 \\
    Ours (Loss, Full)  & \underline{84.2} & 82.6 & \textbf{95.7} & \underline{89.1} & 93.2 & 76.3 & 93.1 & 73.5 & 92.4 & 68.3 \\
    Ours (C+GT, Random)  & 83.4 & 82.7 & \underline{95.0} & 86.1 & 95.4 & 88.5 & 92.9 & 74.9 & \textcolor{red}{\textbf{94.4}} & \textbf{85.5} \\
    Ours (Loss, Random)  & \textbf{84.8} & \textbf{84.6} & \textbf{95.7} & \textbf{90.8} & \textbf{95.8} & \textbf{94.3} & \underline{94.0} & \underline{77.7} & \underline{93.3} & 79.4 \\
    Ours (C+GT, Rejection)  & 83.3 & \underline{83.0} & 94.9 & 86.3 & \underline{95.3} & \underline{91.2} & 93.5 & 76.3 & \textcolor{red}{\textbf{94.4}} & \textcolor{red}{\textbf{86.1}} \\
    Ours (Loss, Rejection)  & \textcolor{red}{\textbf{86.7}} & \textcolor{red}{\textbf{87.1}} & \textcolor{red}{\textbf{95.9}} & \textcolor{red}{\textbf{91.1}} & \textcolor{red}{\textbf{96.2}} & \textcolor{red}{\textbf{94.9}} & \textbf{94.1} & \textbf{77.8} & \textbf{93.5} & \underline{82.0} \\
    \bottomrule
\end{tabular}
\vspace{-0.45cm}
\label{tab:comparions_representations}
\end{table*}

\myparagraph{Setup for data \& model independent attacks.} For independent attacks, we employ different segmentation models for shadow models and victims, as summarized in Table~\ref{tab:data_model_overview}.
Particularly, \bdd has completely compatible label space to \cityscapes of 19 classes, but \mapillary has 65 labels. We merge the some classes from \mapillary into \cityscapes, and ignore the others. For victim models, we train a \pspnet and an \upernet using the official split of \cityscapes, leading to 79.7 and 76.6 mIoU for segmentation.  For shadow models, we apply our splits with balanced In/Out distribution to train a binary classifier. In the end, the F-score of random guess for image-level independent attack is 63.13\%. Comparing it to the numbers in Table~\ref{tab:comparions_representations}, we observe all the attackers obtain much higher F-score than 63.13\%, which shows the severe information leakage of semantic segmentation models.

\myparagraph{Results.}
Results of the different versions of our model as well as comparision to previous work in presented in Table~\ref{tab:comparions_representations}.
While previous work on membership inference targets classification models~\cite{salem2019ml,white_vs_black}, we facilitate a comparison to these approaches by extending them to the segmentation scenario. 
\cite{salem2019ml} proposes a learning-based attacker and a learning-free attacker.
We train their learning-based attacker with 1$\times$1 vector inputs, and test on all pixel locations. Final image-level attacks are obtained by averaging the binary classification scores of all locations. Similar to our method, we test different settings, and it fails to achieve attacks with the shadow model \dpc~\cite{chen2018dpc_nips} in Table~\ref{tab:data_model_overview}. Besides, we test their learning-free attacker by averaging the confidence scores of all locations. Equally, we facilitate a comparison to \cite{white_vs_black} where we use the loss map for the segmentation output.
For our methods, we report the numbers for last two patch selection strategies with sampling 10 patches.
Besides, we also perform attacks with full image inputs using our binary classifiers, which have a global average pooling in the end and are able to handle different sizes of inputs.
We emphasize that the ratio of In/Out testing examples are different for dependent and independent attacks, therefore, the numbers between them cannot be compared.
We conclude that recent models for semantic segmentation are susceptible to membership inference attacks with AUC scores of the attacker up to 87.1\% in the dependent and 94.9\% in the independent setting.
Overall, we observe that our loss-based method with rejection scheme performs best in most settings and measures. 

\begin{figure*}[!t]
\begin{center}
\begin{tabular}{@{}r@{}r@{}r@{}r@{}r}
   \includegraphics[width=0.2\linewidth]{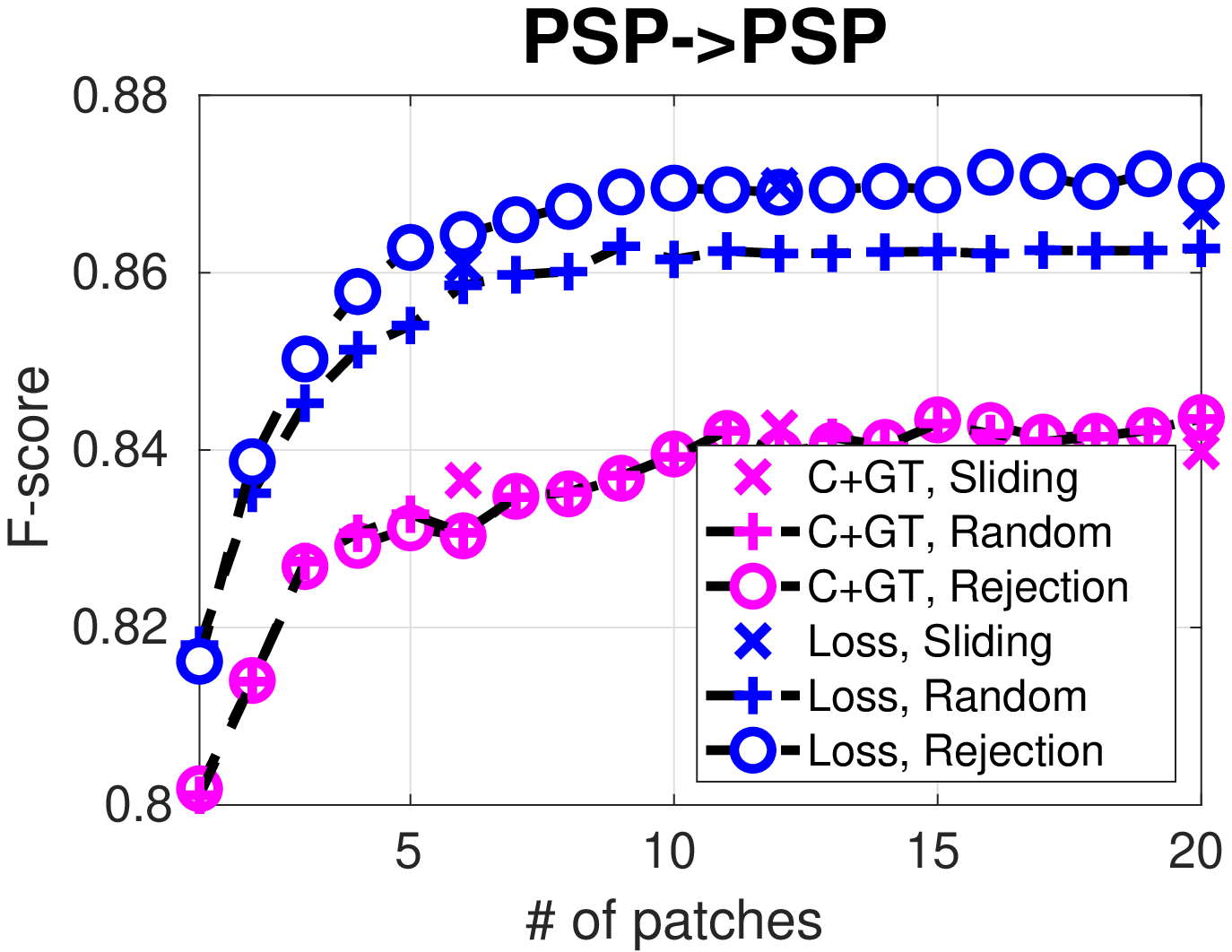} &
   \includegraphics[width=0.1975\linewidth]{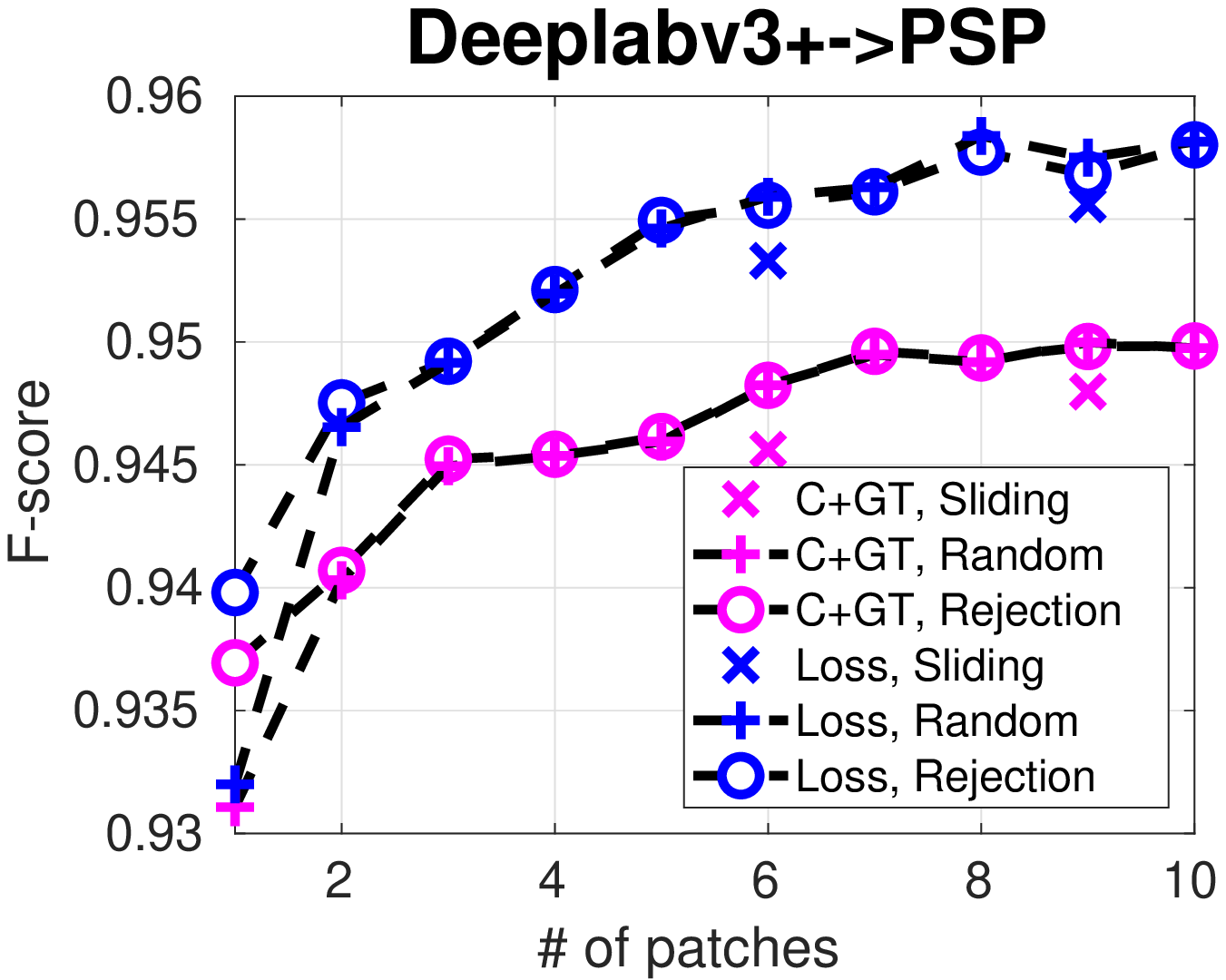} &
   \includegraphics[width=0.198\linewidth]{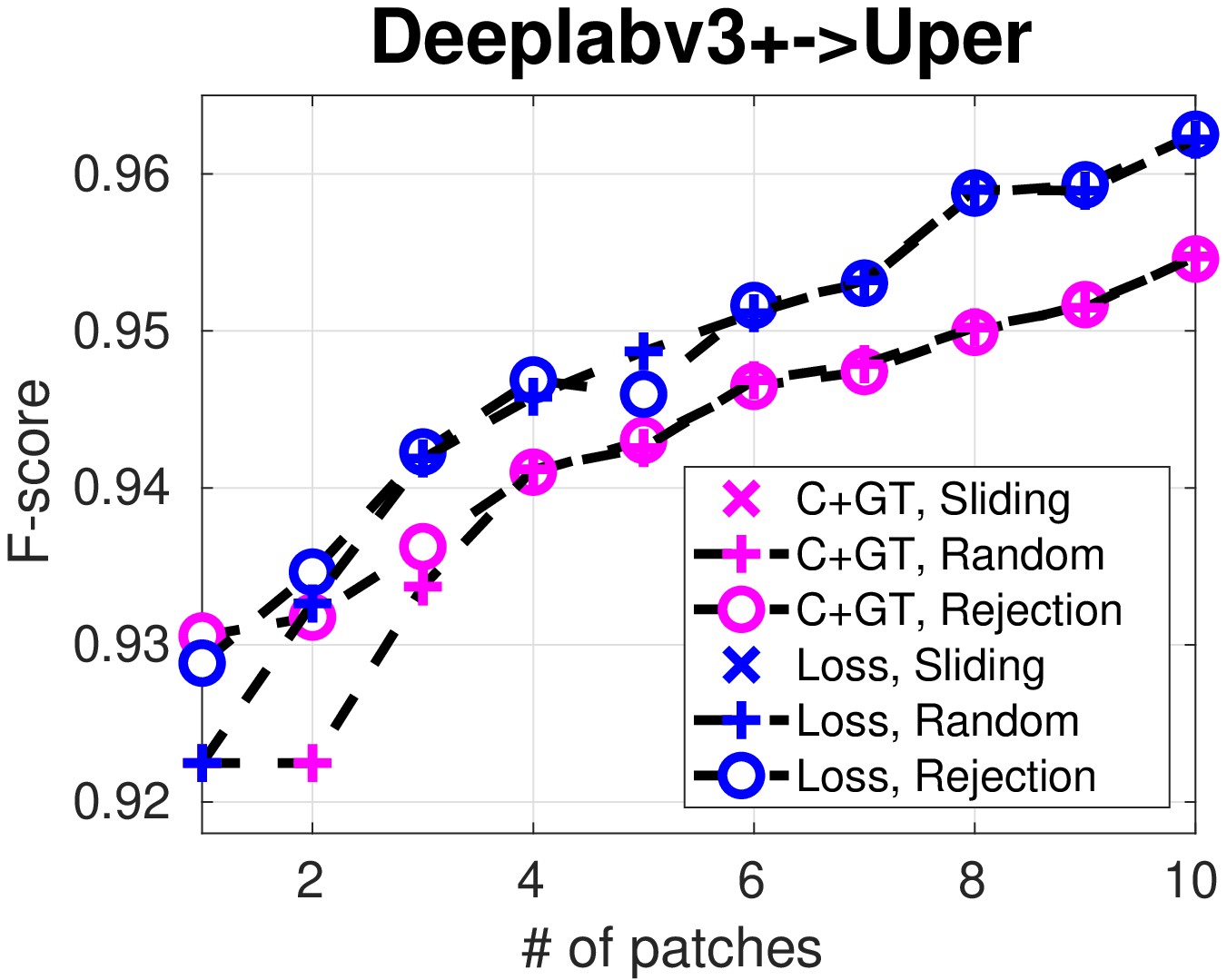} &
   \includegraphics[width=0.198\linewidth]{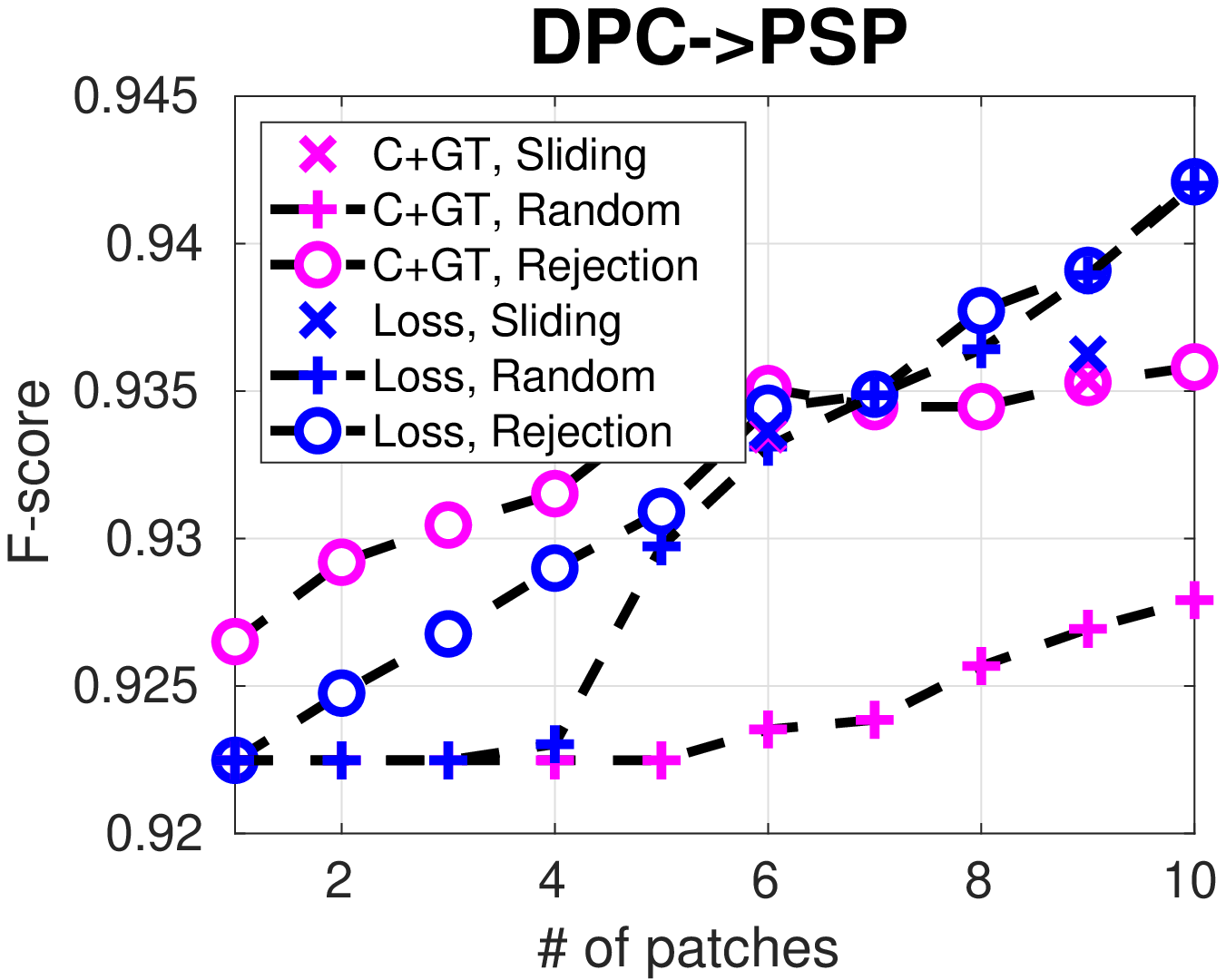} &
   \includegraphics[width=0.2075\linewidth]{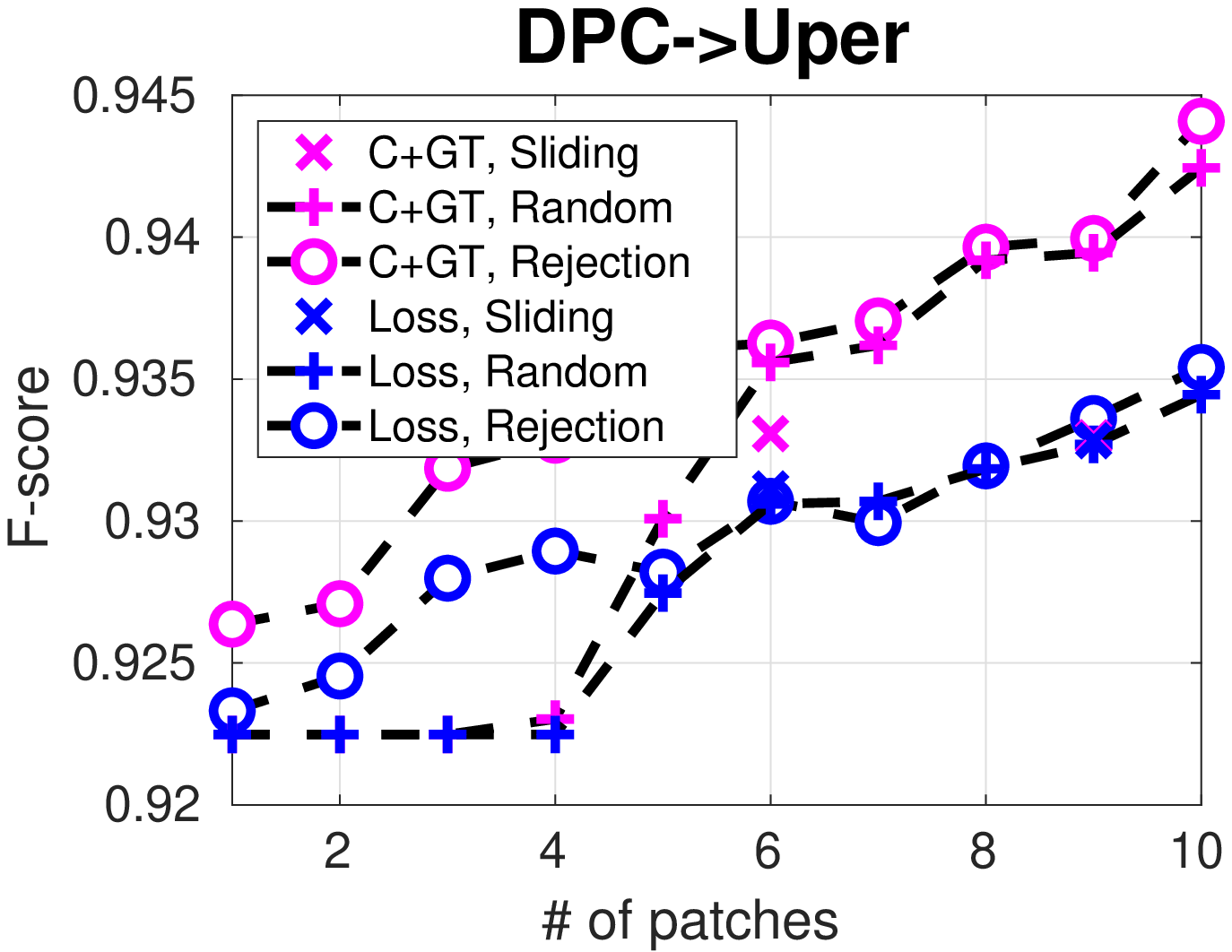} \\
   \includegraphics[width=0.198\linewidth]{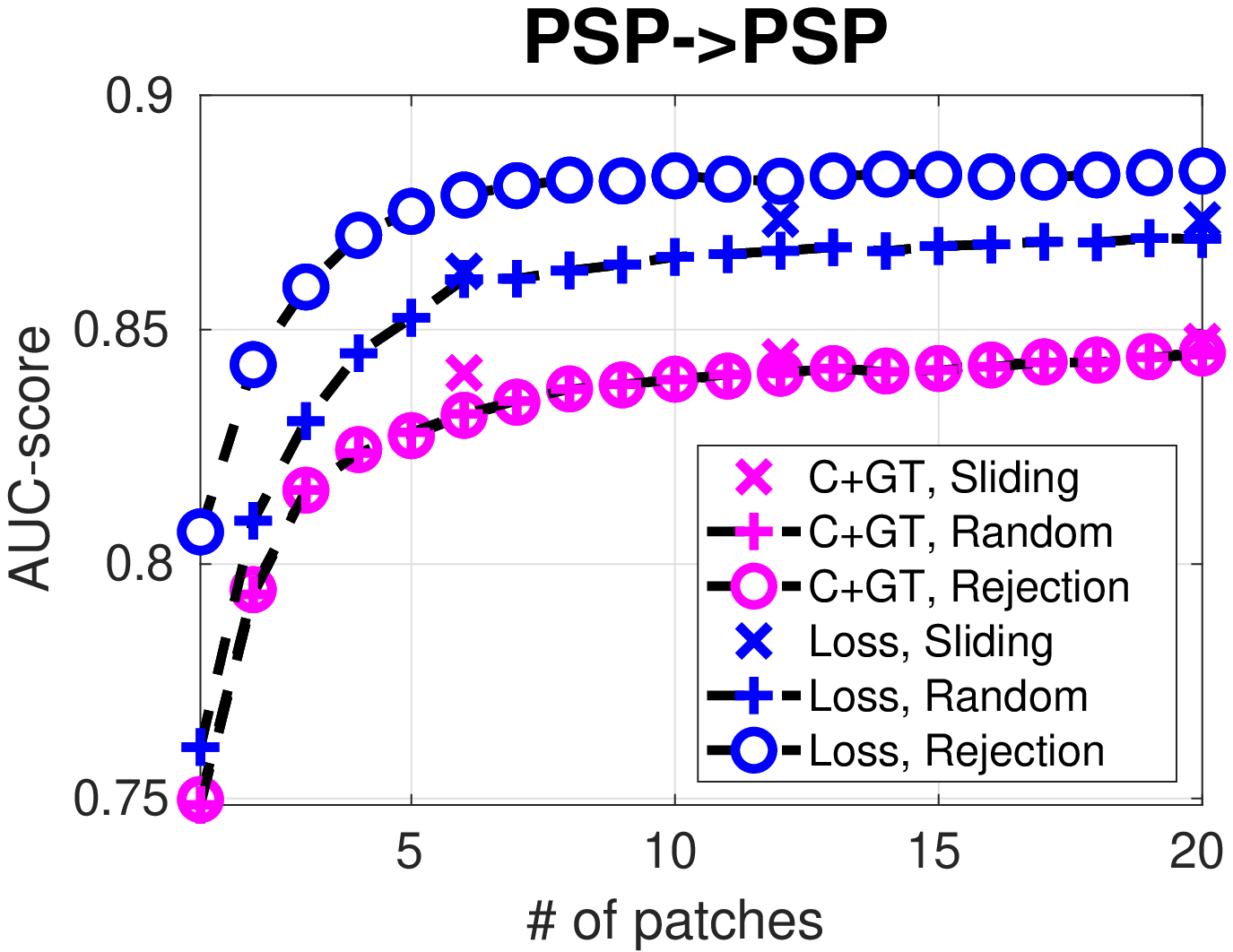} &
   \includegraphics[width=0.198\linewidth]{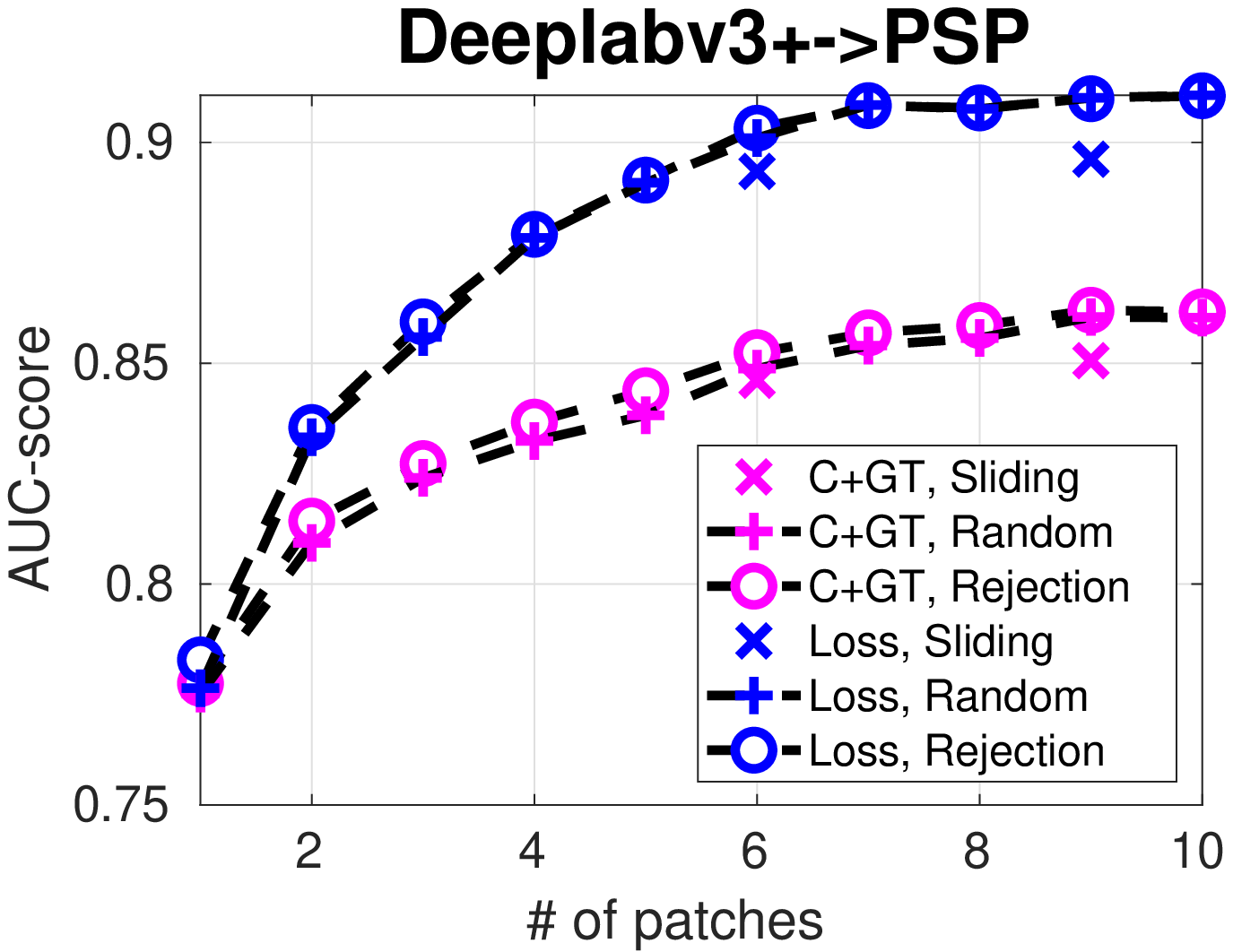} &
   \includegraphics[width=0.198\linewidth]{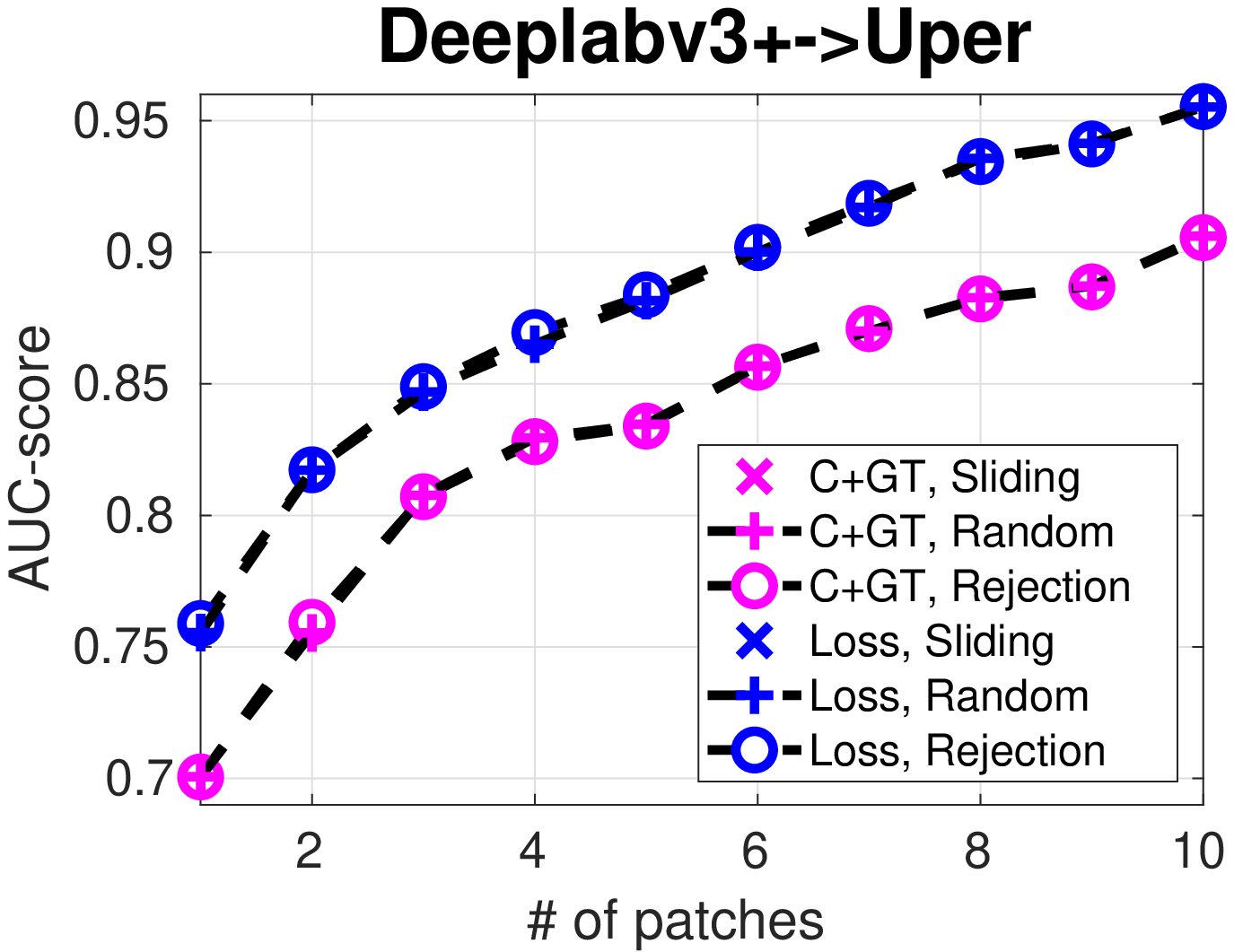} &
   \includegraphics[width=0.198\linewidth]{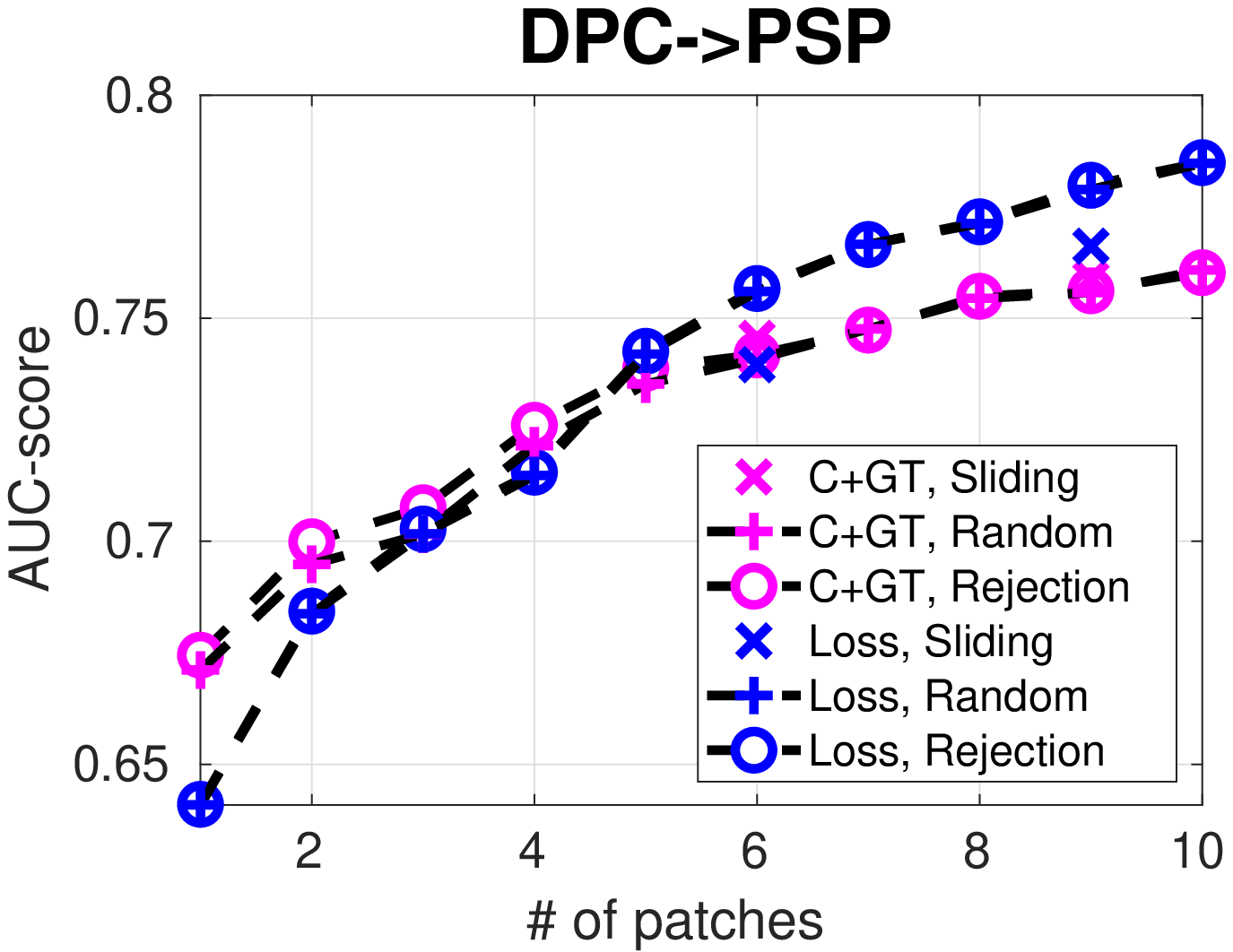} &
   \includegraphics[width=0.199\linewidth]{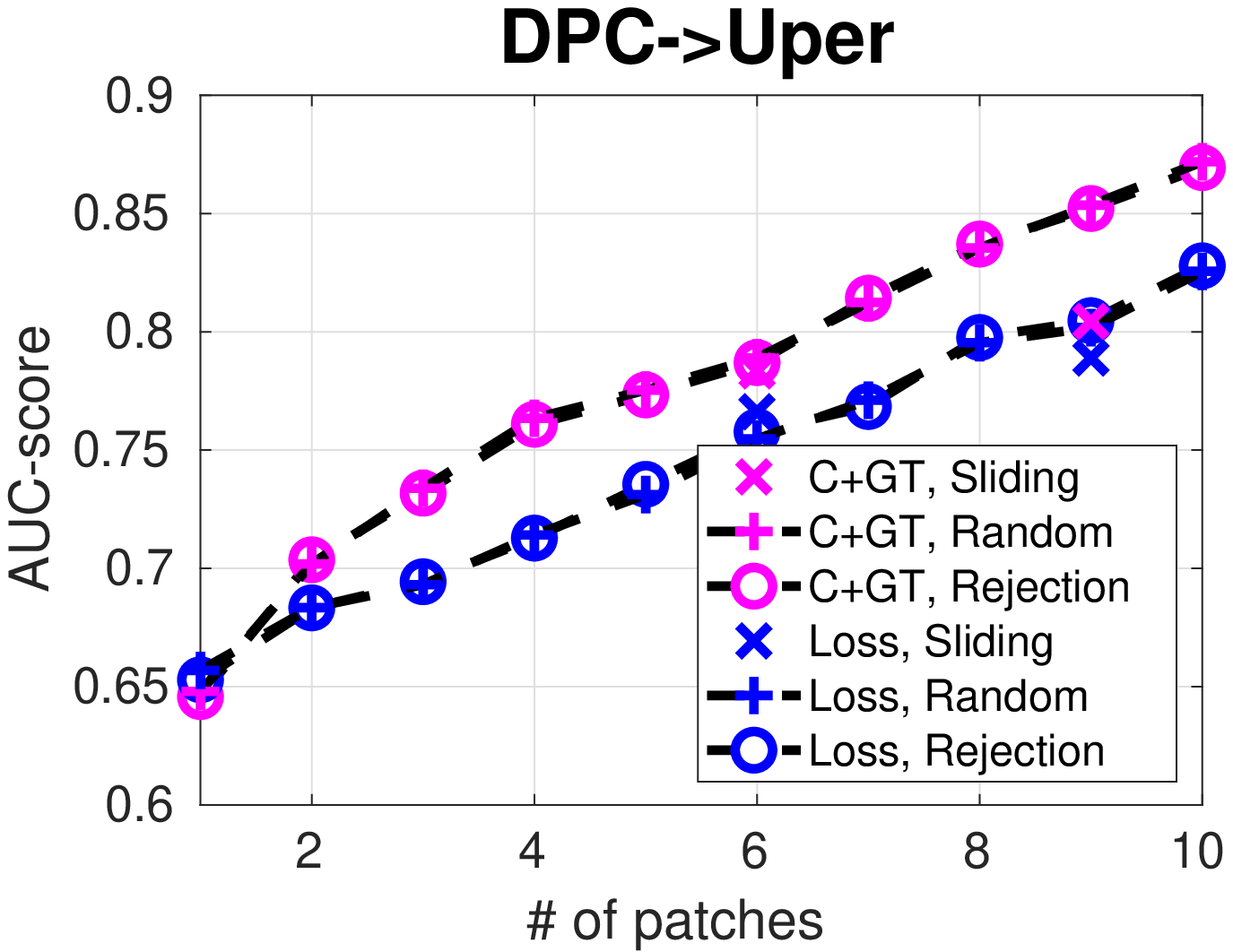}
   \end{tabular}
   \end{center}
   \vspace{-0.4cm}
   \caption{Image-level comparison results w.r.t. {\bf patch selection} and {\bf data representations}, under varying patch numbers.}
   \vspace{-0.7cm}   
\label{fig:ablation_patch_selection}
\end{figure*}

\vspace{-0.06cm}
\myparagraph{Importance of spatial structures.}
Key to strong attacks is exploiting the structural information of an output from a segmentation model.
Hence, we conduct attacks with gradually reduced structural inputs in our dependent attacks in order to  analyze the importance of this structural information for our goal. 
Our final model takes 90$\times$90 blocks as inputs for per-patch attackers.
Therefore, we crop sub-blocks from our final model with input sizes of 60, 45, 30, 15 for providing different level of structures.
We compare the precision-recall curves for per-patch attacks and ROC curves for image-level attacks in Fig.~\ref{fig:ablation_patch}. 
We note that all the feature vectors in the blocks of different sizes have the same scale of receptive fields.
We apply the same architecture of per-patch attacker for sizes 90-30, but modify the ResNet-50 with fewer pooling operations for size 15, because its spatial size is too small.
First, we compare the per-patch attack performance, and are able to observe that attacks become harder with decreasing patch sizes, where smaller patches provide less structures. Second, we compare image-level attacks for them, where random selection strategy is applied to integrate all patches. We sample 5, 20, 20, 30, 30 random patches for size 90, 60, 45, 30, and 15 to integrate image-level results.
Consequently, size 90 achieves the best performance, even though other attackers obtain very close image-level results. Last, we highlight that our concatenation-based attacker degenerates to previous work~\cite{salem2019ml} with 1$\times$1 vector inputs. We observe that 1$\times$1 inputs keep this decreasing trend and achieve worse results than size 15, which can be found in Table~\ref{tab:comparions_representations}.
From this results, we conclude that structures are of great importance in membership inference attacks for semantic segmentation, so that an attacker is able to mine some In/Out confidence or loss patterns over an array input.

\vspace{-0.15cm}
\myparagraph{Analysis of patch selection and data representation.}
We test our three sampling strategies and two representations as depicted in section~\ref{subsec:methods}. Fig.~\ref{fig:ablation_patch_selection} plots the image-level comparison results.
For sliding windows, we sample at least 6 patches to guarantee an entire image can be covered. For random locations, we sample different numbers of patches for image-level attacks to observe the influence of patch numbers, starting from one patch. We conduct this experiments for 3 times and report the mean. In summary, we observe these two strategies achieves comparable performance when the same numbers of patches are used.
Specifically, sliding windows perform better on dependent attacks with loss maps, and random locations are better for independent attacks (\deeplab$\xrightarrow{}$PSP, and \deeplab$\xrightarrow{}$Uper), which may be caused by inconsistent data distributions or different behaviors of segmentation models.
Last, we test our random locations with rejection strategy. To avoid the affect of random seeds, we sample the same locations to previous random locations if a patch is not rejected. We can see clear improvements if we sample very few patches, whose results are sensitive to sampled locations. In street scenes, road has a large portion of pixels, therefore, it tend to sample a road patch, which has the highest accuracy over all the classes and less discrimination for In/Out classification. After ignoring those patches, performance is improved because the rejection helps us avoid those less informative patches. To conclude, not all the regions contribute to successful attacks for segmentation, that we need a regime to determine membership status of an image, instead of processing the whole like previous work for classification.

Comparing our patch-based attacks to the full image attacks, we realize using full images as inputs makes performance significantly decreased, even though the same classifier is applied. The classification for full images may be affected by misleading areas. Hence, partitioning an image into many patches helps focus on local patterns and makes a better decision. Besides, we observe that our rejection scheme achieves better performance than random scheme, which further supports our argument on the difference between segmentation and classification. In addition, our concatenation-based attacker outperforms ~\cite{salem2019ml}, which demonstrates the importance of spatial structures, similar to Fig.~\ref{fig:ablation_patch}.
From our results, \cite{white_vs_black} is able to obtain acceptable performance but worse than our structured loss map-based attackers, which hold the structural information.  Finally, our novel structured loss maps achieve better results than concatenation and other methods~\cite{salem2019ml,white_vs_black} in most cases.

\section{Defenses}
\label{sec:defenses}
\vspace{0.12cm}
To mitigate the membership leakage and protect the authority of a model, we study several defenses for semantic segmentation models, while maintaining their utility with little performance degeneration. When a model is deployed, a service provider has all rights to access the model and data. Our work shows for the first time a feasible solution for protecting very large semantic segmentation model. As a consequence, we manipulate the model in training or testing stages by reducing the distribution gaps between training data and others w.r.t. confidence scores of predictions or loss values, including Argmax, Gauss, Dropout and DPSGD.
The first two methods can be applied in any segmentation models and last two can be applied in deep neural networks. 

\myparagraph{Settings.}
We analyze the performance of image-level attacks according to random locations in this section, which are easily compared to the results without defenses in Table~\ref{tab:comparions_representations}. Because Gaussian noises, or dropout will change output distributions, rejection scheme may sample different patches,  we only test the random location schemes and sample patches at the same locations for different defenses, and keep consistent to previous attacks.

For dependent attacks, our shadow and victim models have the same post-processing and learning protocol, as claimed in section~\ref{sec:settings}. In other words, we employ the same defenses and strength factors for them in this setting.
For independent attacks, we report the settings of \deeplab$\xrightarrow{}$PSP and \deeplab$\xrightarrow{}$Uper, which are the most successful. We only employ defenses on victim models as protections for released black-box semantic segmentation models.

\myparagraph{Evaluation methodology.}
Due to the different ratios for In/Out examples in various settings, we do not report their F-scores in this section. Instead, we only employ AUC-score to compare different defense methods, expecting to reduce the original attacks' AUC-scores to 0.5, that random guess in all the settings hold this number.
Furthermore, an ideal defense is supposed to make attacks hard and preserve segmentation utility at the same time.
Therefore, we apply mIoU~\cite{long2015fully} to evaluate the segmentation performance and jointly compare different defense methods w.r.t. capability of membership protection and utility of segmentation.

\subsection{Methods and Results}
\begin{figure}[!h]
\begin{center}
\begin{tabular}{@{}r@{}r@{}r}
   \includegraphics[height=2.25cm,width=2.475cm]{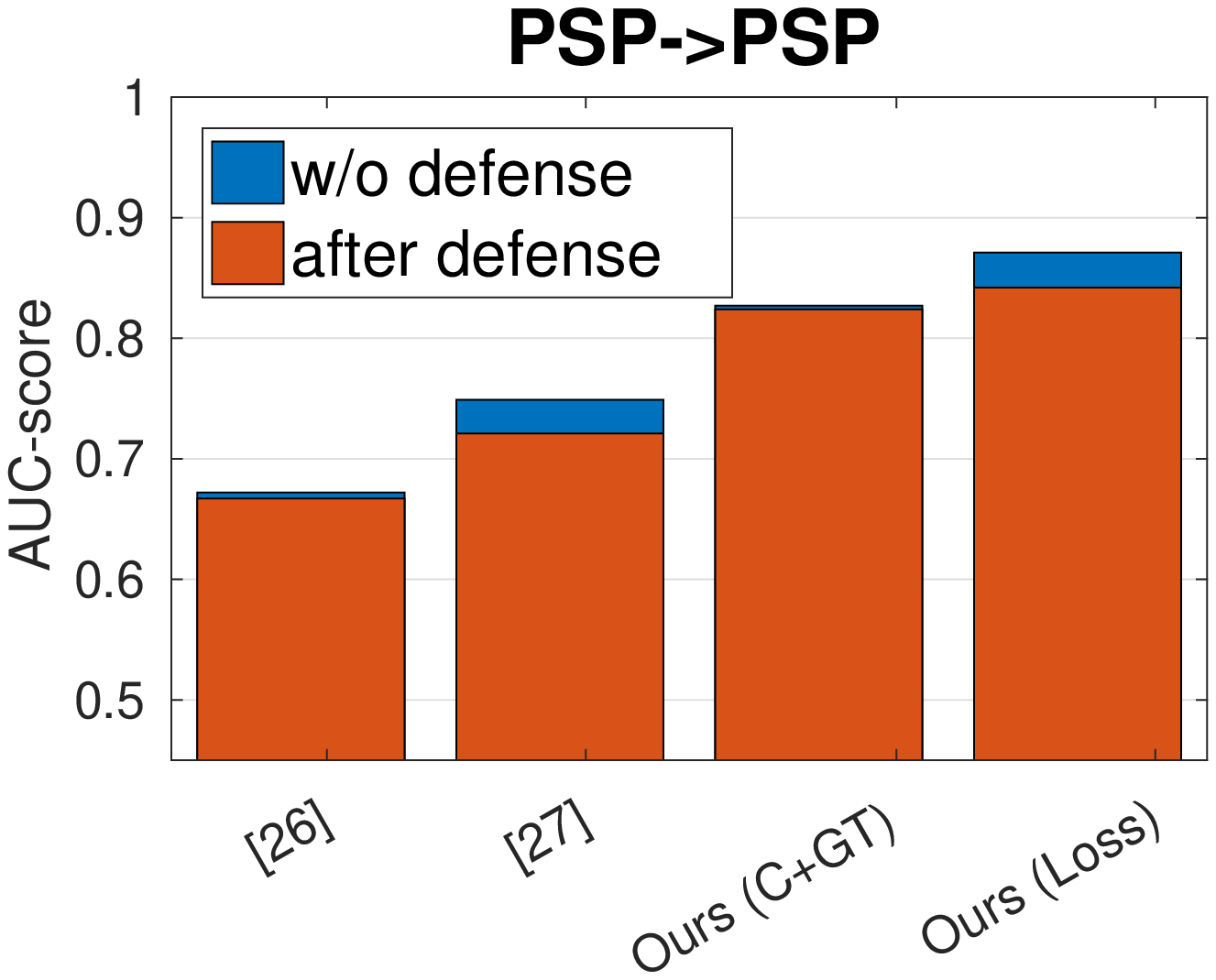} &
   \includegraphics[height=2.25cm,width=2.475cm]{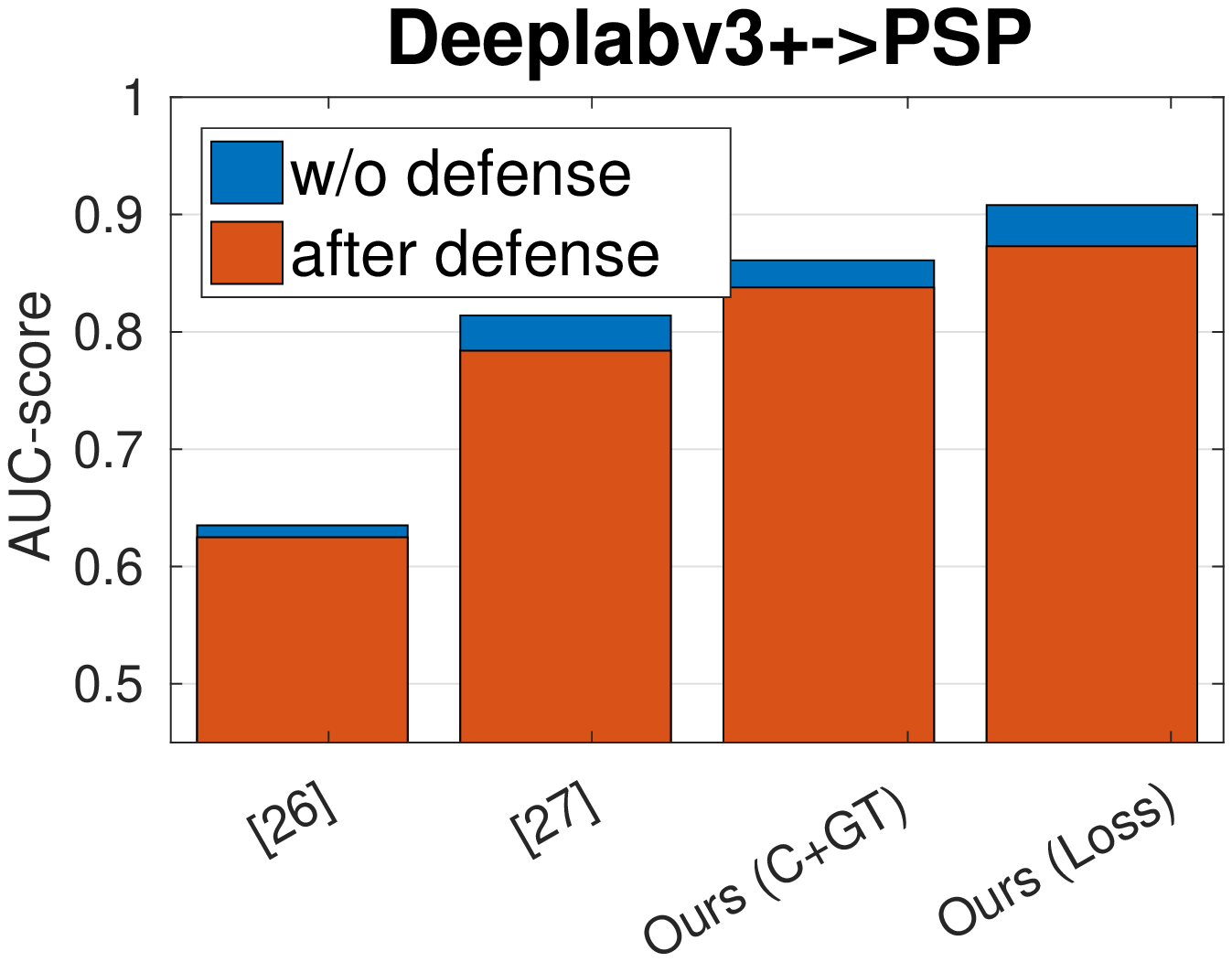} &
   \includegraphics[height=2.25cm,width=2.475cm]{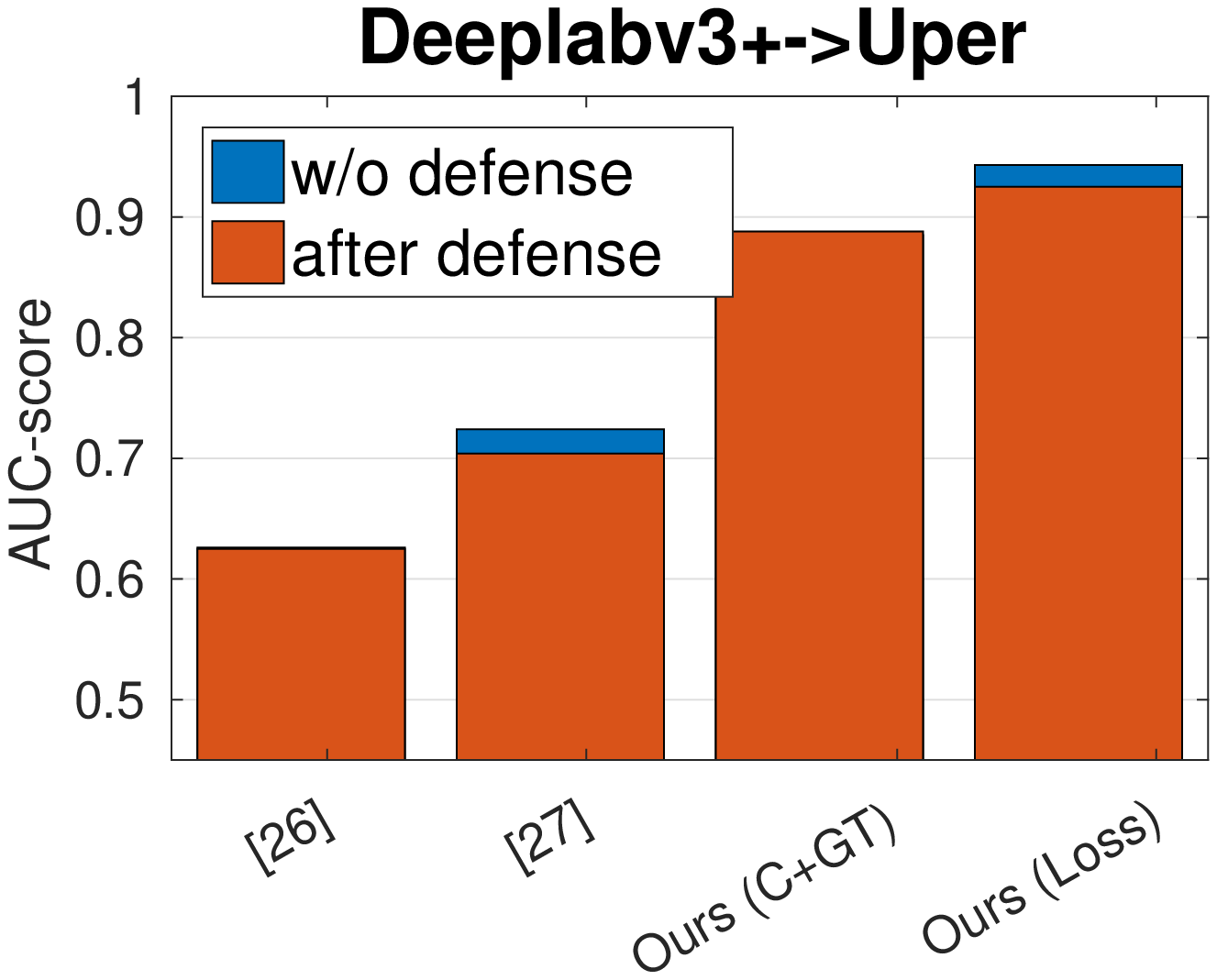}
   \end{tabular}
   \end{center}
   \vspace{-0.6cm}
       \caption{Performance comparison for {Argmax defense}.}
          \vspace{-0.7cm}
\label{fig:defense_argmax}
\end{figure}

\vspace{-0.2cm}
\myparagraph{Argmax.} It only returns predicted labels instead of posteriors for an image. We use one-hot vectors to complete attacks for our methods and others~\cite{salem2019ml,white_vs_black}.
Obviously, previous learning-free attacker~\cite{salem2019ml} based on confidence scores fails to recognize membership states, because every example has confidence 1. 
In Fig.~\ref{fig:defense_argmax}, we show the comparison results for all the other methods.
Because argmax is very easy to be noticed, we train binary classifiers for independent attacks with argmax operation as well.
In general, argmax only reduces membership leakage in segmentation models a little for all the attackers.
A model already leaks information when it only returns predicted labels. 
To conclude, we highlight the difference to protecting classification, that argmax cannot successfully protect the membership privacy for segmentation.

\begin{wrapfigure}{r}{0.66\textwidth}
\vspace{-0.65cm} 
\centering
\scalebox{1}{
\begin{tabular}{ccc}
\includegraphics[height=2cm,width=2.475cm]{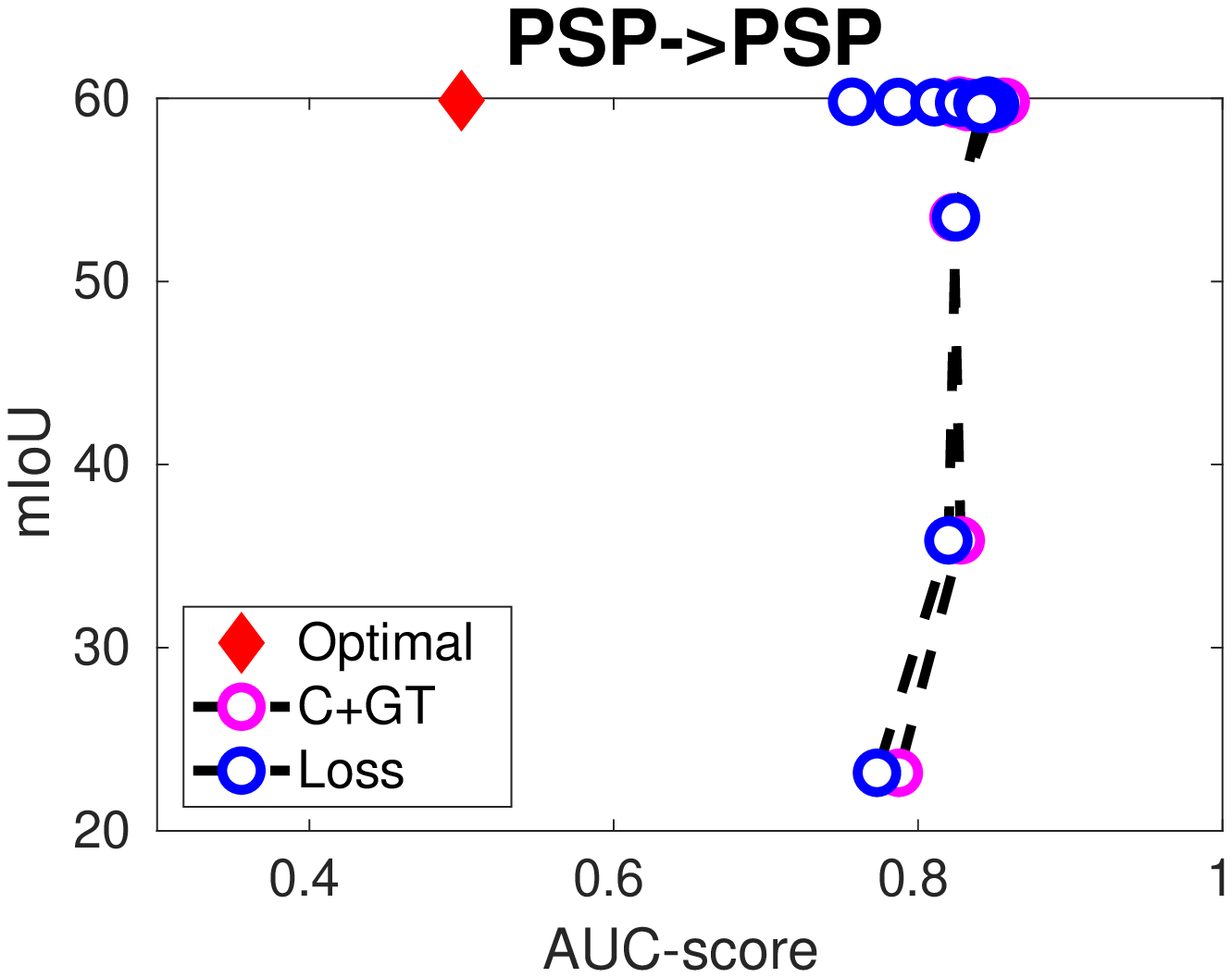}&
\includegraphics[height=2cm,width=2.475cm]{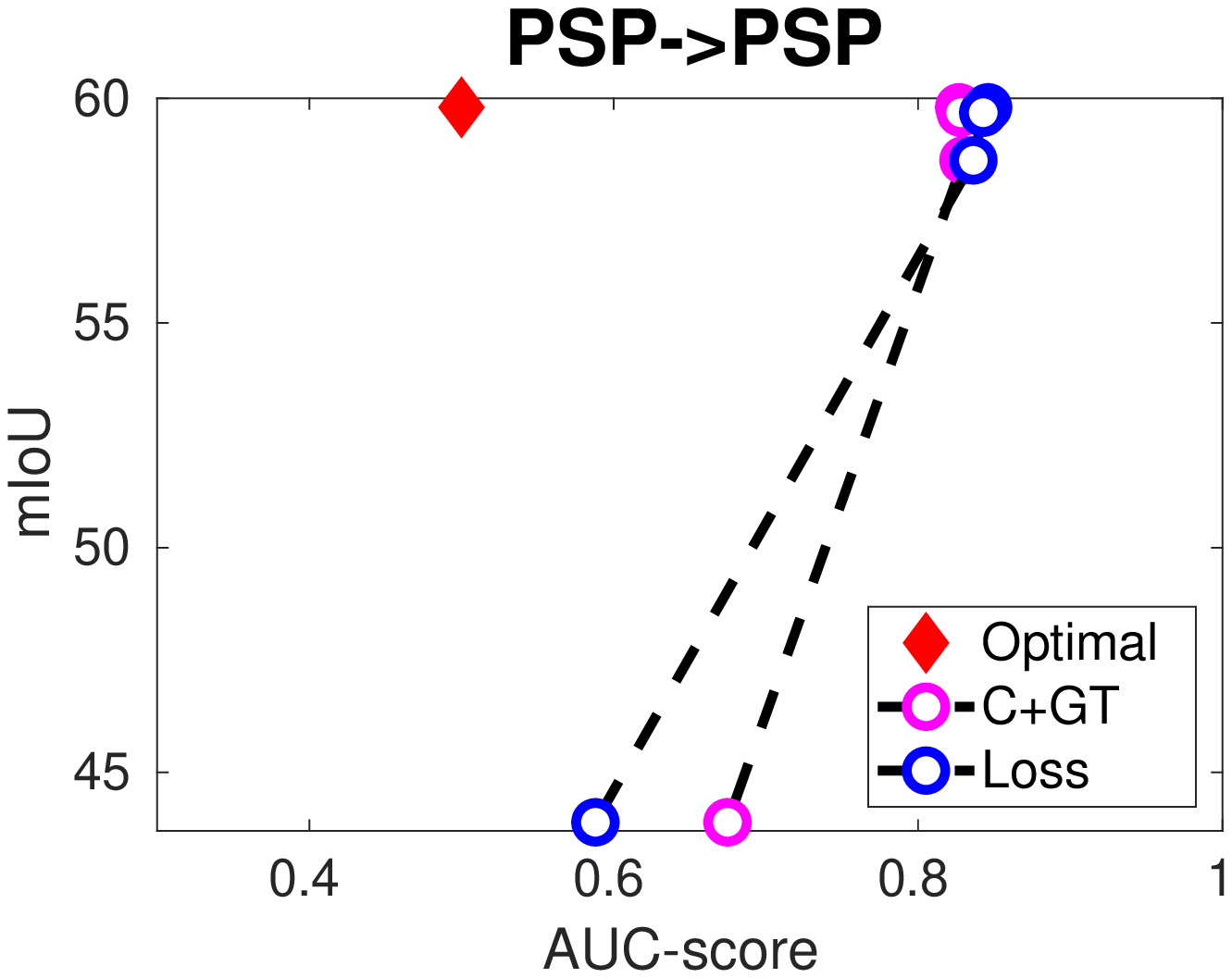}&
\includegraphics[height=2cm,width=2.475cm]{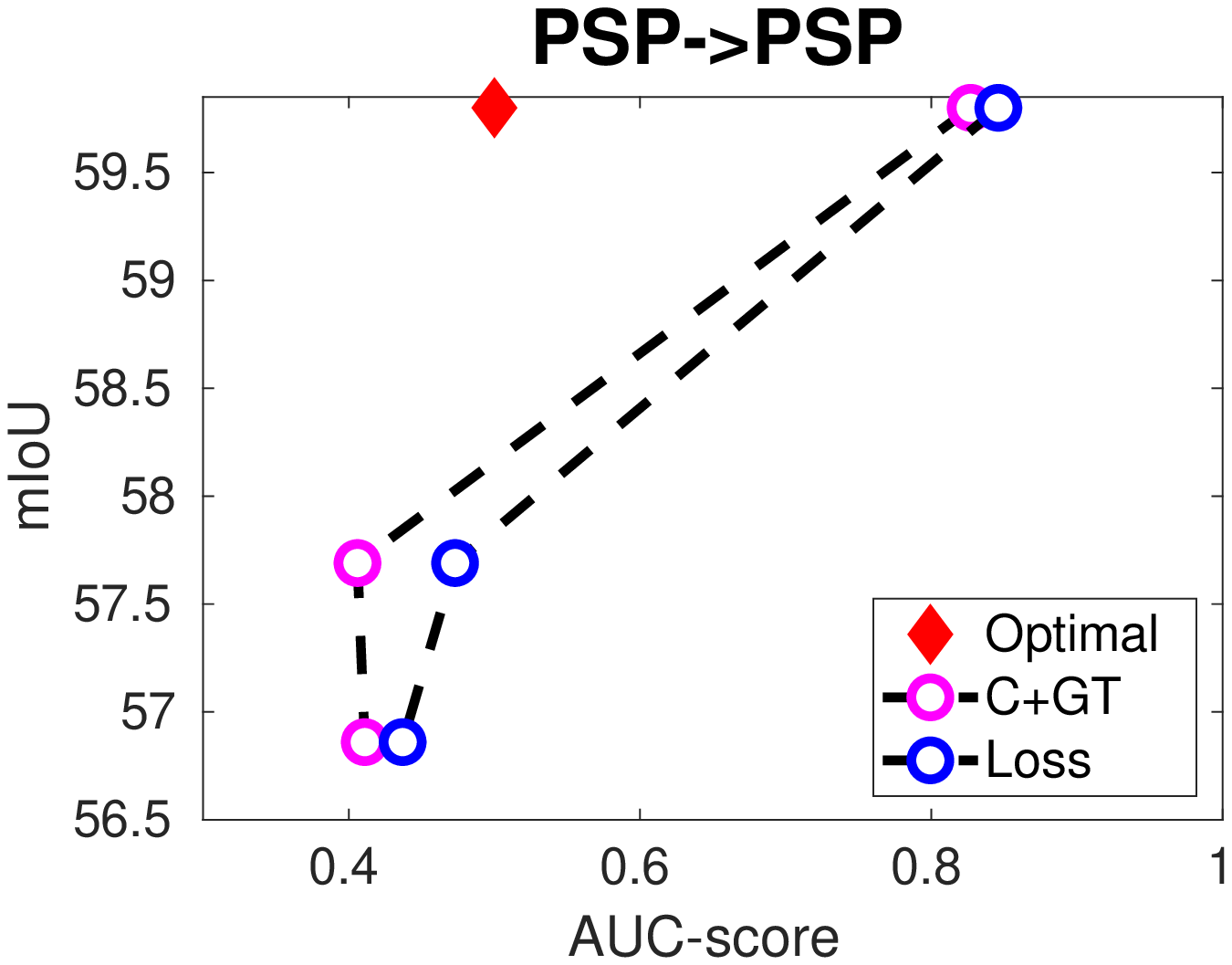}\\
\includegraphics[height=2cm,width=2.665cm]{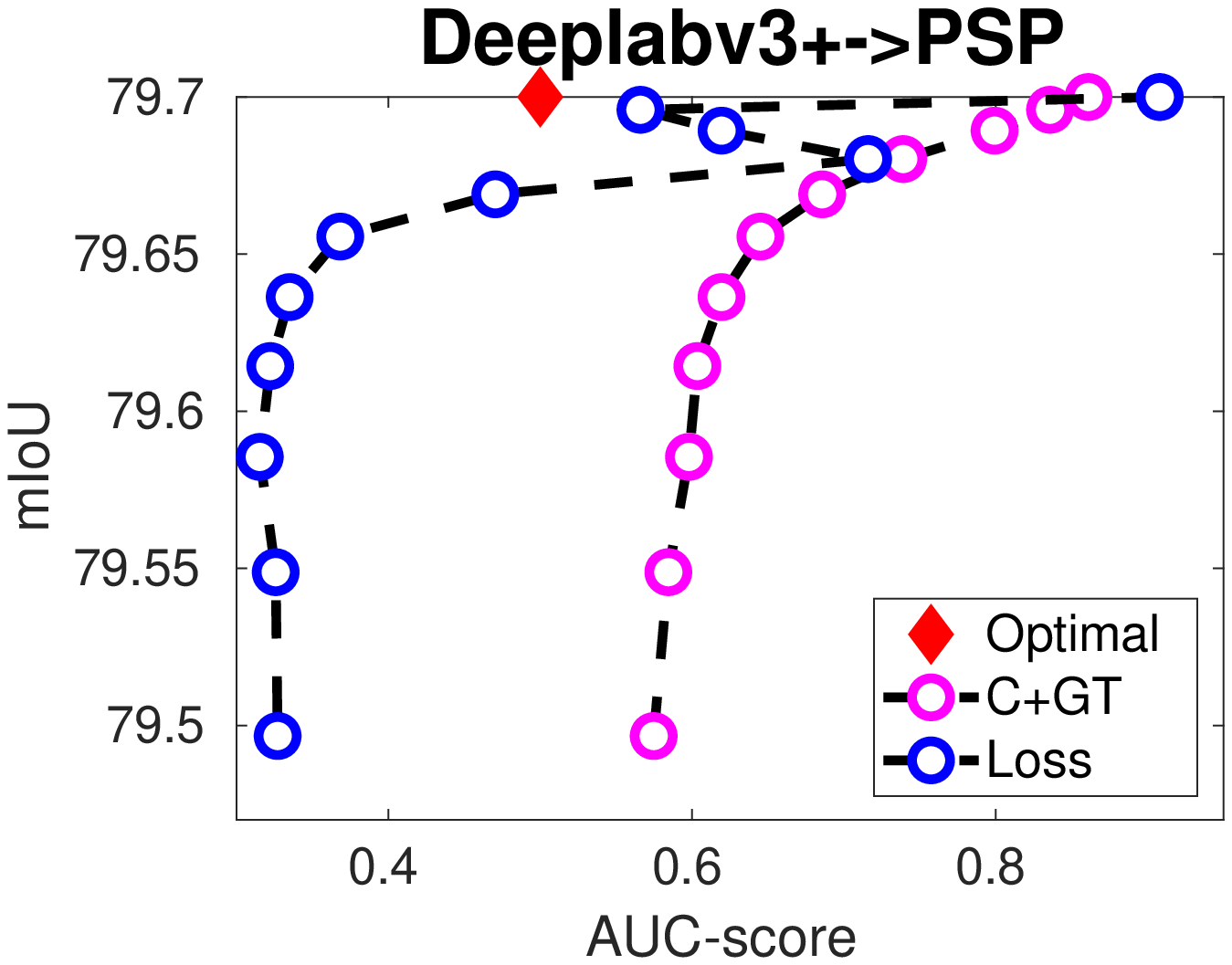}&
\includegraphics[height=2cm,width=2.665cm]{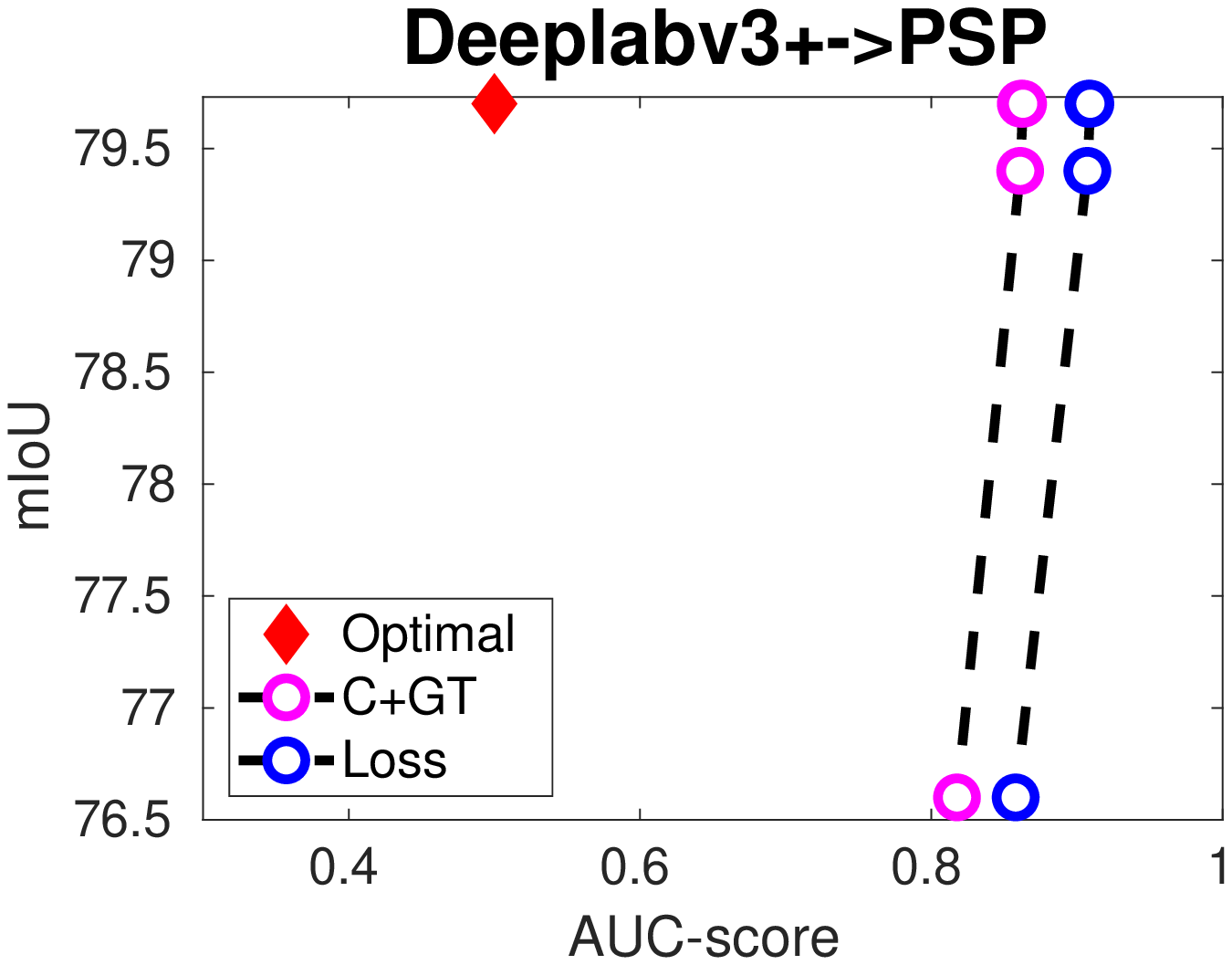}&
\includegraphics[height=2cm,width=2.565cm]{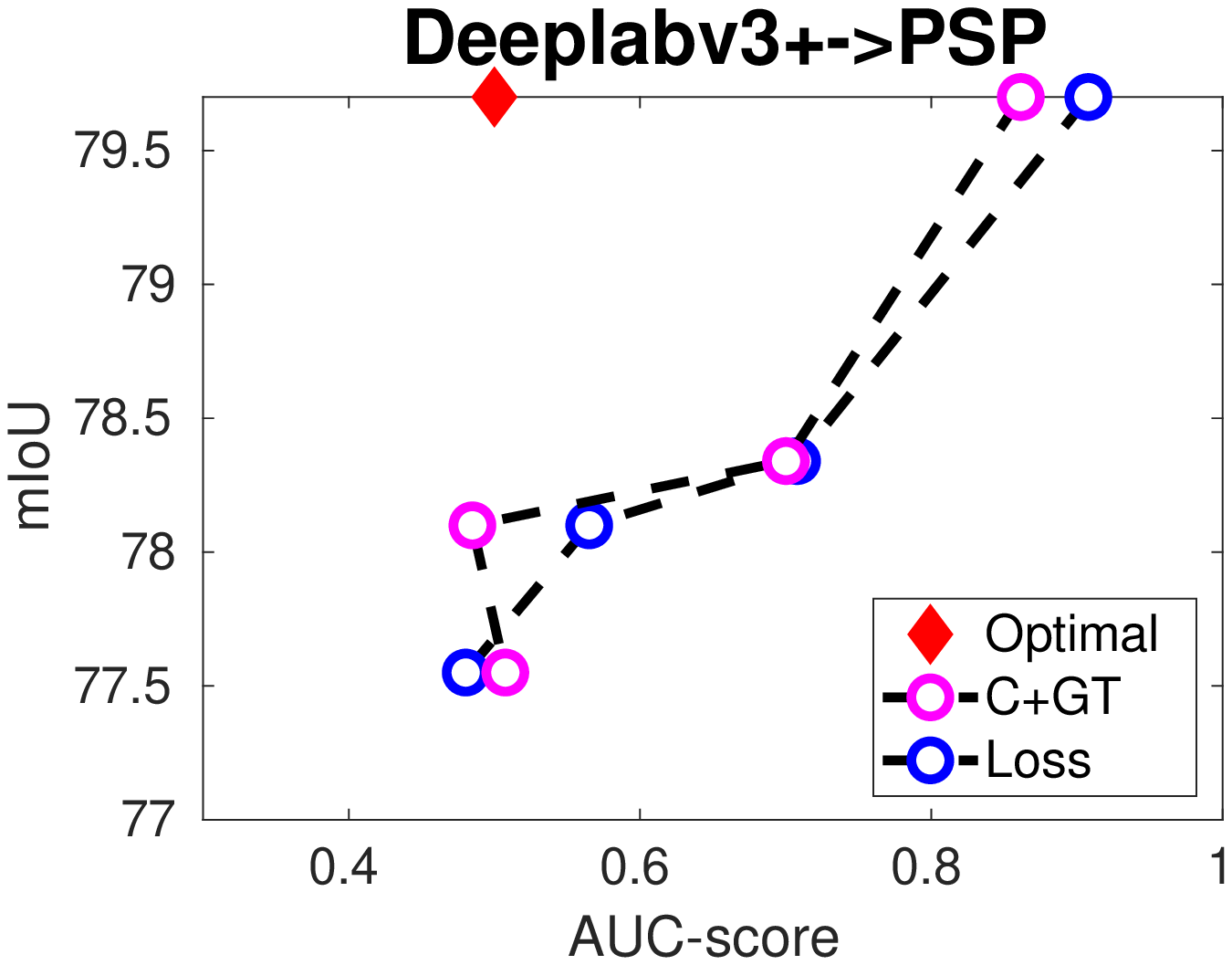}\\
\includegraphics[height=2cm,width=2.665cm]{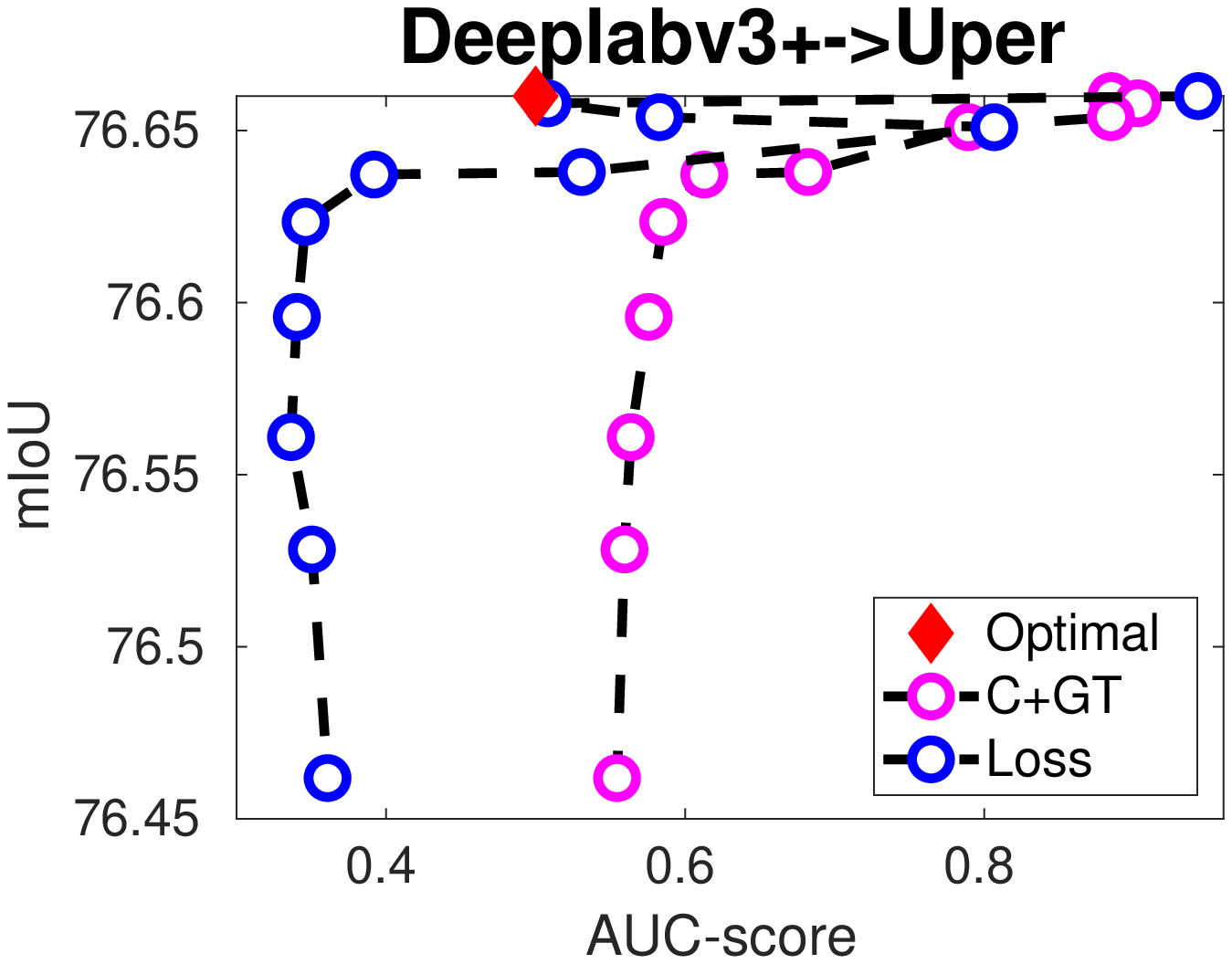}&
\includegraphics[height=2cm,width=2.475cm]{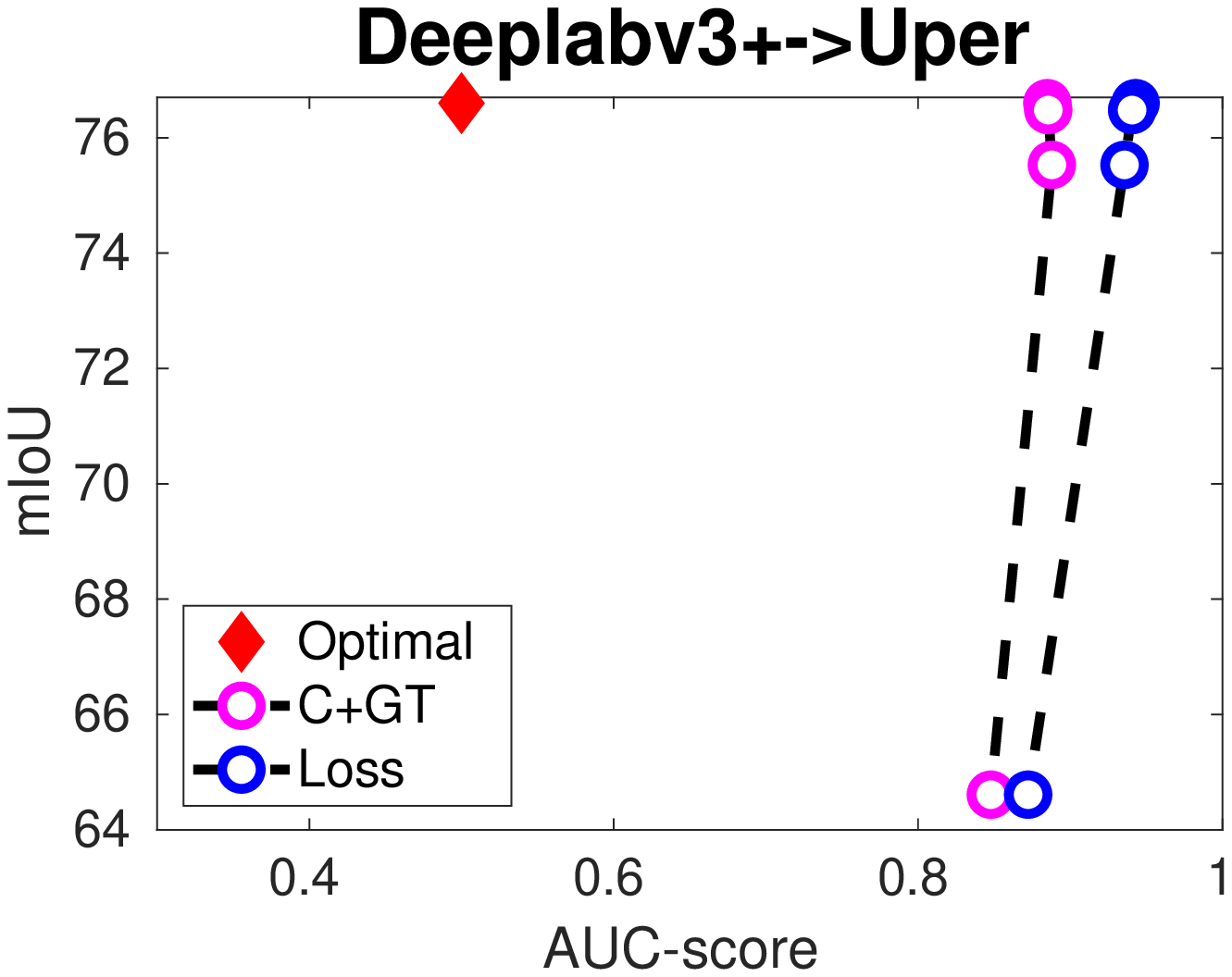}&
\includegraphics[height=2cm,width=2.475cm]{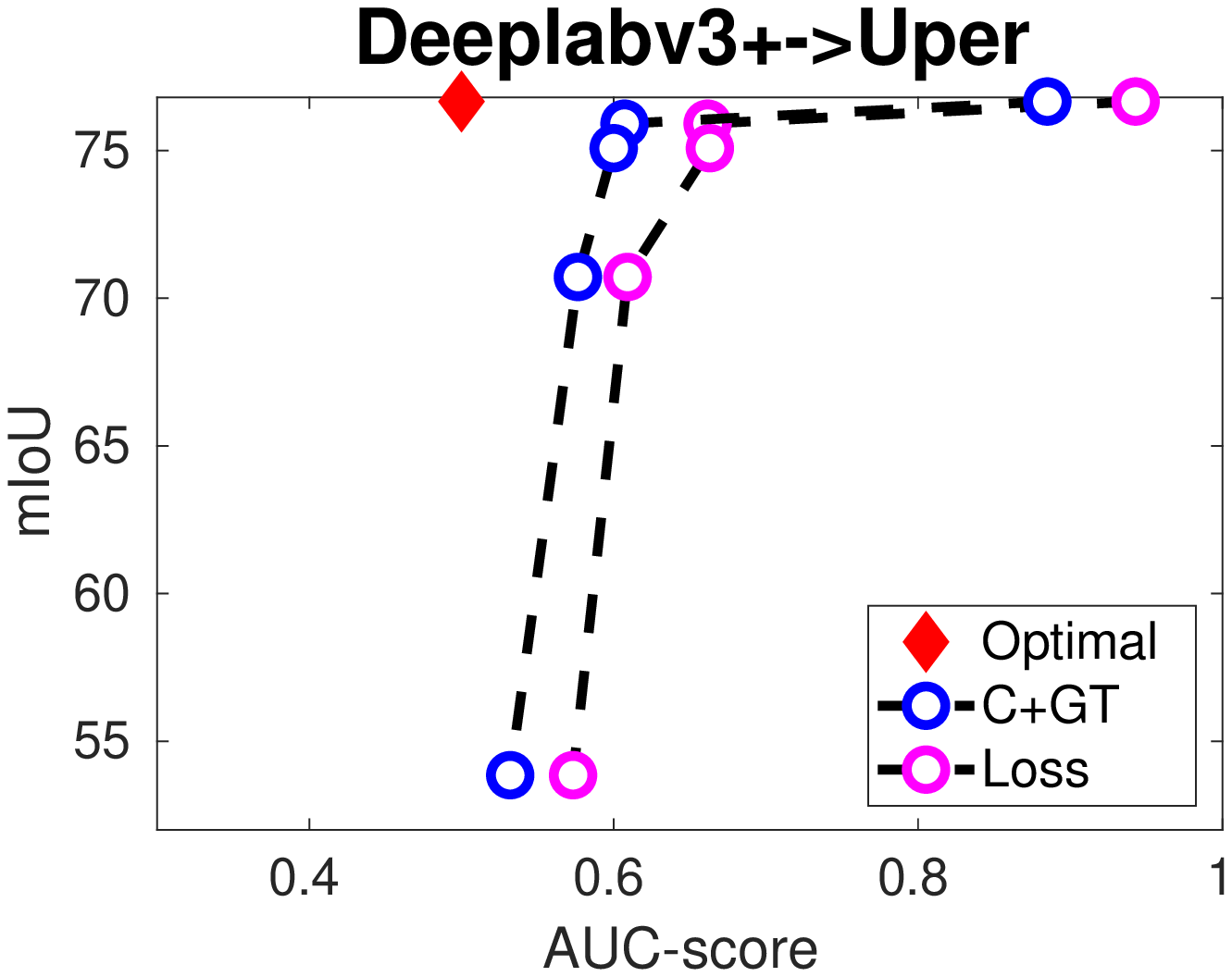}\\
(a) Gauss &
(b) Dropout &
(c) DPSGD 
\end{tabular}
}
\vspace{-0.15cm} 
\caption{ Joint plots for different defenses. $x$-axis is the AUC-score for membership protection and $y$-axis is the mIoU for segmentation utility. optimal defenses achieve 0.5 AUC-score while preserving segmentation utility, as drawn with the red diamond.
}
\vspace{-0.45cm} 
\label{fig:defense_overall}
\end{wrapfigure}

\myparagraph{Gauss.} To hide overfitting artifacts or patterns, we add Gaussian noises on the posteriors with different variances, varying from 0.01 to 0.1 with step 0.01 for independent attacks. 
To further test the defense for dependent attacks, we add very strong noises to variance 0.4.  
After noising, we set the values into 0 in case they are smaller than 0, and then normalize each location individually.
Segmentation performance is decreased with stronger noises, therefore, we show the joint privacy-segmentation plots in Fig.~\ref{fig:defense_overall}(a) to observe the defense behaviors as well as the maintained utility of the segmentation method. First, we observe Gauss protects PSPNet and UperNet in independent attacks successfully, which reduces AUC-scores from ~0.9 to less than 0.6, while only losing 0.2 mIoU. Second, we observe our loss-based attackers are more sensitive to Gaussian noises. Despite stronger attacks of structured loss maps, they are easier to protect with Gaussian noises.
Finally, we realize this defense is hard to mitigate leakage for dependent attacks. Even though we employ very strong noises for this, losing mIoU from 59.88 to 23.17, it still has more than 0.75 AUC-scores for both attacks. To conclude, Gauss is hard to protect a model when the noises of the same distribution are added to victim and shadow models, and binary classifiers can pick useful information from noisy inputs.

\myparagraph{Dropout.} It is used to avoid overfitting in training a deep neural networks, that we applied in training our victim model with dropout ratio 0.1. However, it does not hide membership from our studies in section~\ref{sec:attacks}. Therefore, we enable dropout operation during testing to blur a prediction. 
We realize a network still produces decent results when we use a different dropout ratio. Hence, we apply dropout ratio 0.1, 0.5 and 0.9 to obfuscate a prediction at different degrees.
We show the joint plots in Fig.~\ref{fig:defense_overall}(b). From our study, we observe enabling dropout during test is able to slightly mitigate membership leakage, but segmentation performance decreases a lot when a large ratio is applied.

\myparagraph{DPSGD.} Differential Privacy SGD (DPSGD)~\cite{abadi2016deep} adds Gaussian noises on the clipped gradients for individual examples of a training batch, in a way that the learnt parameters and hence all derived results such as predictions are differentially private.
We apply DPSGD in our study to protect a model. Before training, we collect gradient statistics over entire training data for different layers of a network, and set individual clipping factors for all the layers. Next, we train PSPNet with Gaussian variances 10$^{\text{-3}}$, 4$\times$10$^{\text{-3}}$ for dependent settings, and variances 10$^{\text{-3}}$, 4$\times$10$^{\text{-3}}$, 8$\times$10$^{\text{-3}}$ for independent settings. For UperNet, we train with 10$^{\text{-6}}$, 10$^{\text{-3}}$, 3$\times$10$^{\text{-3}}$, 6$\times$10$^{\text{-3}}$.
Theoretically, the Gaussian noises used in our model is not enough to guarantee a tight differential privacy bound~\cite{dwork2008dp_survey}. However, there is a gap between theoretical
garuantees and emperical defenses. prior work has shown practical
defenses from small gaussian noise and hence loose bounds~\cite{jayaraman2019evaluating}.
In our work, we demonstrate this in semantic segmentation models and show the utility-privacy plots in Fig.~\ref{fig:defense_overall}(c). We observe that DPSGD successfully protect memberships in all the settings, in particular, it will not hamper the utility of segmentation models which only reduces 1.12, 1.36 and 0.75 mIoU when noises with 1e-6 are applied in three rows. Therefore, we recommend DPSGD to train a segmentation model for protecting membership privacy in practise.

\myparagraph{Summary of defenses.}
In spite of the success of membership inference under various settings, we point out feasible solutions which can significantly reduce the risk of information leakage. (1) Adding Gaussian noises helps prevent leakage in independent settings from unknown attackers. Tradeoff between model degeneration and information leakage is able to be considered to choose a suitable noise level.  Hence, we recommend this method as a basic protection without further costs to prevent potential independent attacks, which are very cheap to implement.
(2) For neural networks, we suggest applying DPSGD to train a model, which mitigates the leakage in all the settings with limited model degeneration, even though it adds  noises on the gradients during training and hence requires increased training time.

\vspace{-0.2cm}
\subsection{Interpretability}
\begin{figure}[!t]
\centering
\scalebox{1}{
\centering
\begin{tabular}{@{}l@{\hspace{0.06cm}}c@{\hspace{0.06cm}}c@{\hspace{0.06cm}}c@{\hspace{0.06cm}}c@{\hspace{0.06cm}}c}
\multicolumn{1}{c}{} & \multicolumn{1}{c}{No}&\multicolumn{1}{c}{Argmax}&\multicolumn{1}{c}{Gauss}&\multicolumn{1}{c}{Dropout}&\multicolumn{1}{c}{DPSGD}\\
\vspace{-0.1cm}
\rotatebox{90}{   CAMs}&
\includegraphics[width=0.15\linewidth]{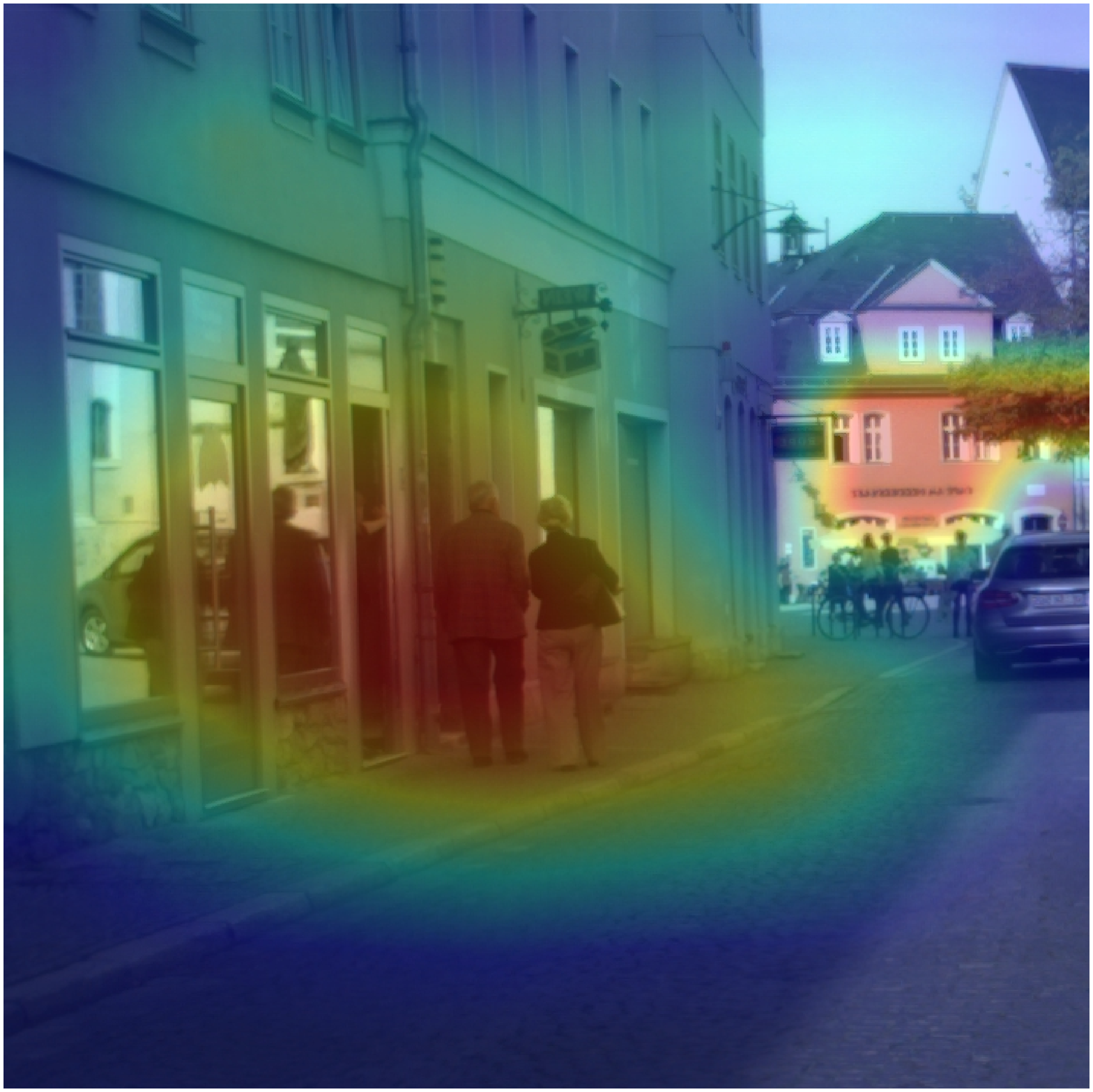}&
\includegraphics[width=0.15\linewidth]{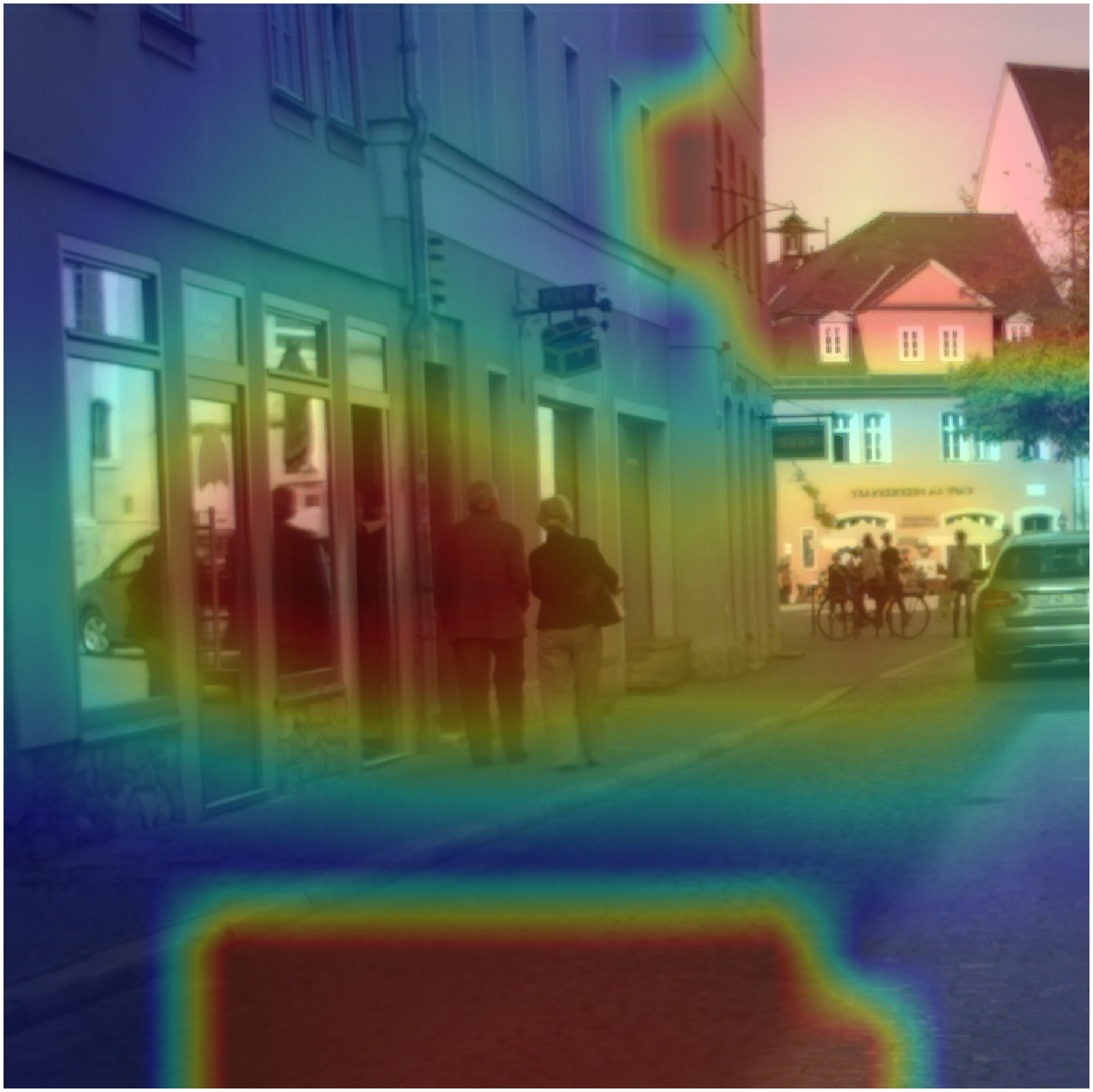}&
\includegraphics[width=0.15\linewidth]{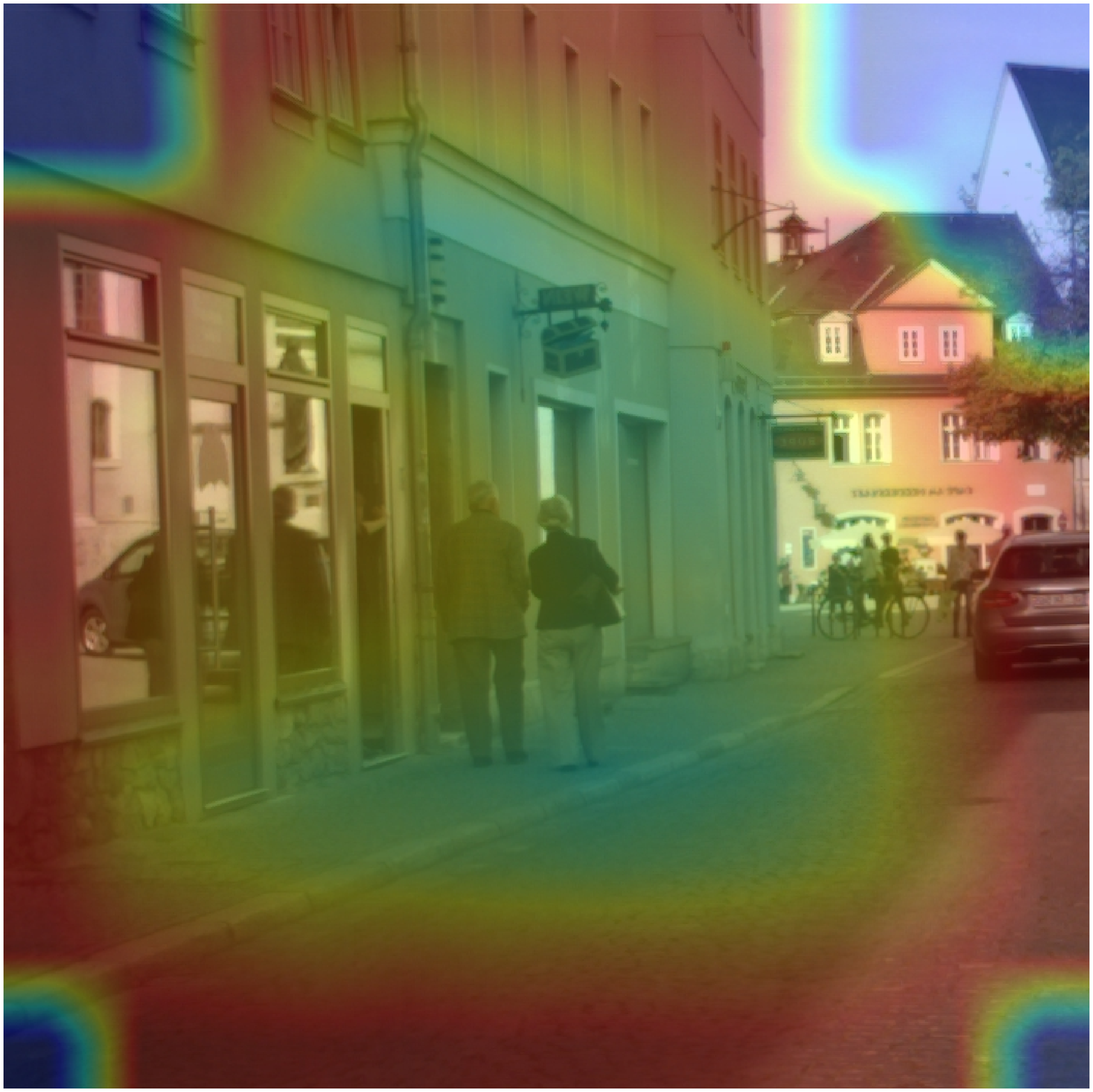}&
\includegraphics[width=0.15\linewidth]{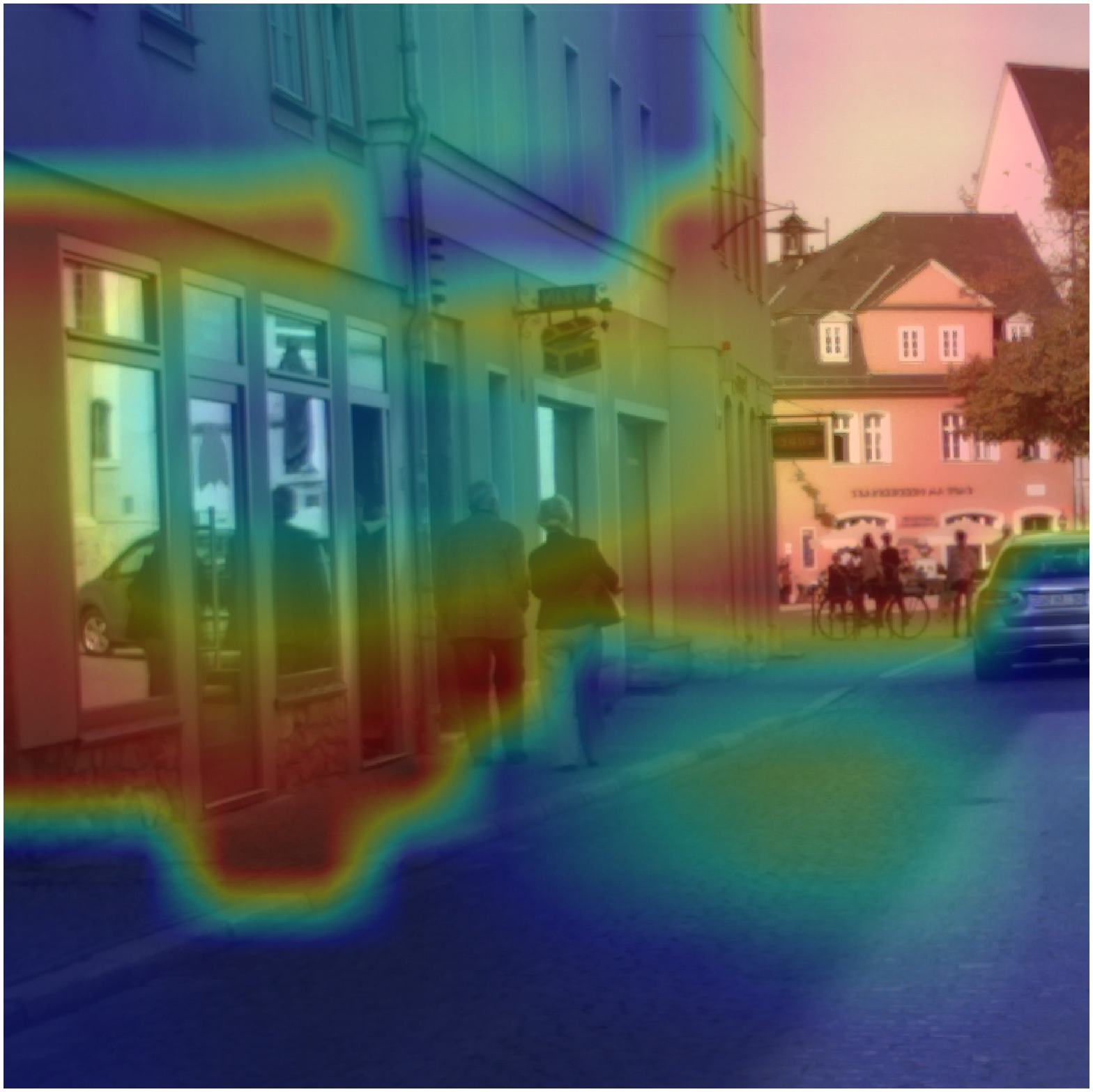}&
\includegraphics[width=0.15\linewidth]{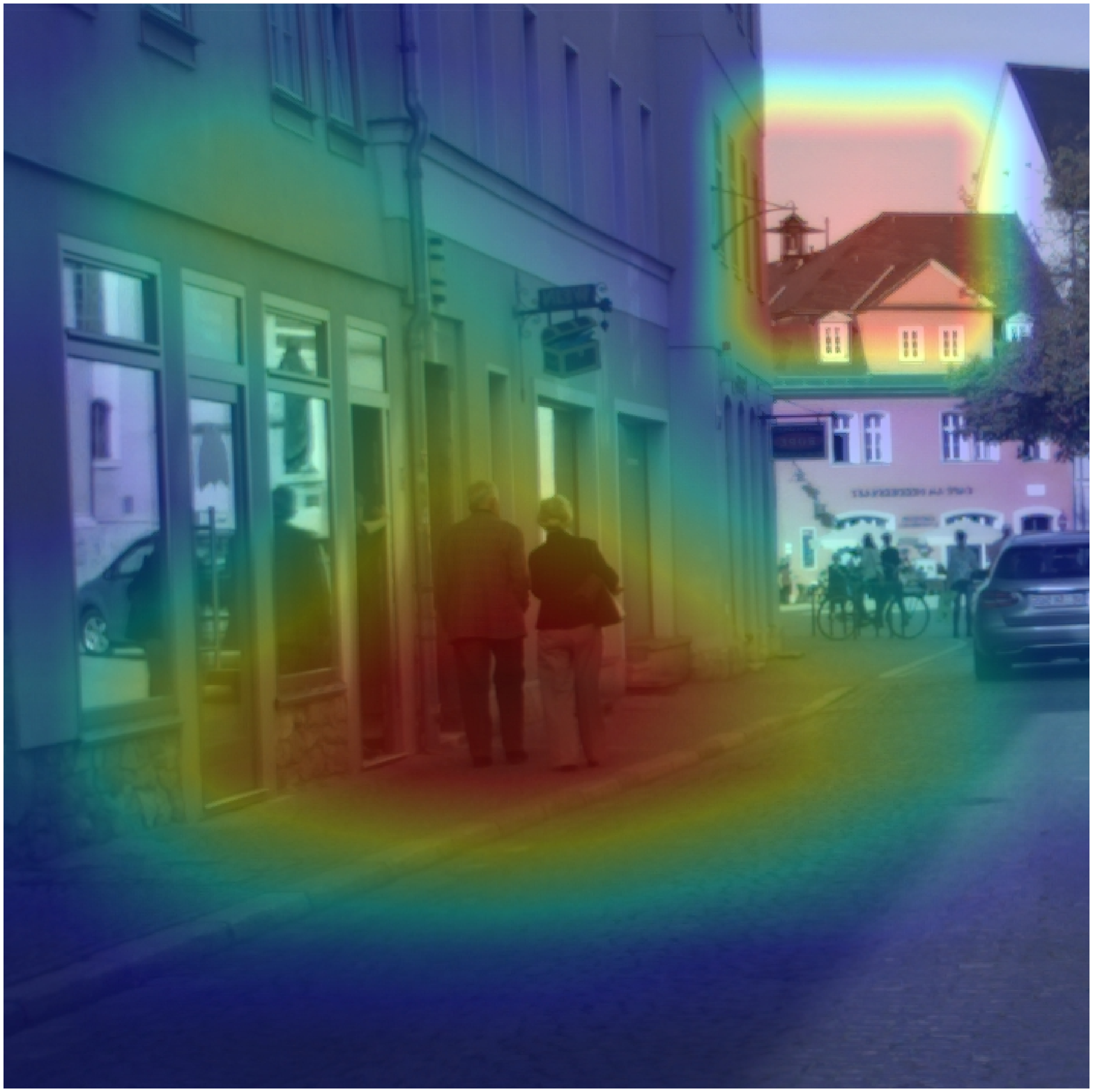}\\
\rotatebox{90}{   SLMs}&
\includegraphics[width=0.15\linewidth]{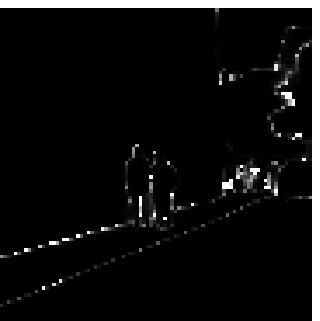}&
\includegraphics[width=0.15\linewidth]{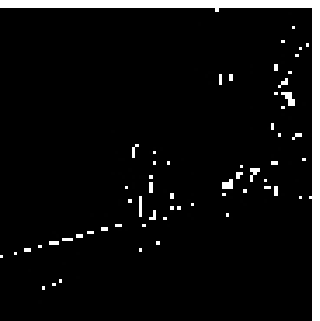}&
\includegraphics[width=0.15\linewidth]{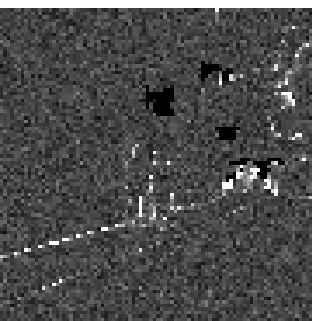}&
\includegraphics[width=0.15\linewidth]{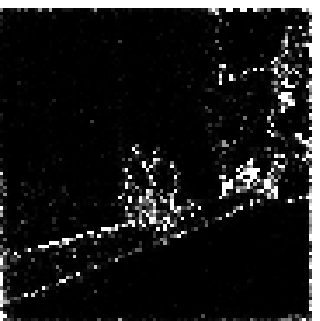}&
\includegraphics[width=0.15\linewidth]{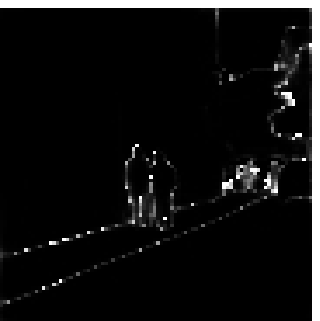}\\
\end{tabular}
}
\caption{Class activation maps (CAMs) and structured loss maps (SLMs) for independent attack \deeplab$\xrightarrow{}$Uper.
}
\vspace{-0.6cm} 
\label{fig:cam_loss_attack}
\end{figure}

One of difference between attacking segmentation and classification is on the input form of the binary classifier, where the input of segmentation can be regarded as an image. Hence, our method can provide interpretations for different examples, indicating important regions for recognizing membership status. Besides, interpretations also help us to understand and compare different defenses. We apply class activation maps (CAMs)~\cite{zhou2016cam_cvpr} to highlight the areas that help to detect examples/patches from training set in Fig.~\ref{fig:cam_loss_attack}. Besides, we also compare the activation areas before and after defenses with structured loss maps. 

First, we observe our attacker is able to mine some regions with specific objects or intersections between two classes, even our attacker has no interaction with a victim. Second, we compare the attacker's different behaviors for those defenses. We can see argmax can simply change the CAM to different intensities, but still hold the major layout of the original CAM.
For Gaussian noises, we employ variance 0.1 here, and can apparently observe noises on the structural loss map for all the pixel locations, therefore, it makes all the examples have a similar CAM. For dropout, it will change structured loss maps in many places and then change the CAM. In particular, it changes the locations with strong loss values more than others. For DPSGD, we can see it has very similar loss maps to the original model. The only differences are on some regions hard to segment. Even DPSGD changes the loss maps a little, the final CAMs are able to change a lot for some examples, therefore, it helps defend stealing memberships while preserving segmentation performance very well.  

\vspace{-0.2cm}
\section{Conclusion}
\label{sec:conc}
\vspace{-0.25cm}
We have provided the first membership inference attacks and defenses for semantic segmentation models by extending previous membership attacker for classification and proposing a new specific representation (i.e., structured loss maps). Our study is conducted under two different settings with various model/data assumptions. We show that spatial structures are important to achieve successful attacks in segmentation, and our structured loss maps achieve the best results among all. Besides, we study defense methods to reduce membership leakage and provide safe segmentation. As a result, we suggest to add Gaussian noises on the posteriors in inference, or apply differential privacy in training. We hope that our work contributes to the awareness of novel threats that modern deep learning models pose -- such as leakage of information on the training data. Our contributions shows that such threats can be mitigated with little impact on the utility of the overall model.

\vspace{-0.2cm}
\subsubsection*{Acknowledgement}
The authors thank Apratim Bhattacharyya, Hui-Po Wang and Zhongjie Yu for their valuable feedback to improve the quality of this manuscript.
\clearpage
%
%
\bibliographystyle{splncs04}
\bibliography{egbib}
\end{document}